\pdfoutput=1

\documentclass[11pt]{article}
\usepackage{amsmath}
\usepackage[final]{acl}
\usepackage{subcaption}
\usepackage{times}
\usepackage{latexsym}
\usepackage{cleveref}
\crefname{figure}{Fig.}{Figs.}
\crefname{table}{Table}{Tables}
\usepackage{array}
\usepackage{tabularx}
\usepackage{ragged2e}  
\usepackage{booktabs}  
\usepackage[T1]{fontenc}
\usepackage{placeins}
\usepackage{enumitem}  
\usepackage{float}
\usepackage[utf8]{inputenc}
\usepackage{makecell}
\usepackage[most]{tcolorbox}  
\tcbuselibrary{breakable,skins}
\usepackage{microtype}

\usepackage{tcolorbox}
\usepackage{alltt}
\tcbuselibrary{breakable}
\definecolor{lightblue}{rgb}{0.85,0.92,1}

\usepackage{inconsolata}
\usepackage{censor}

\usepackage{graphicx}
\usepackage{booktabs}
\usepackage{multicol}
\usepackage{multirow}   
\usepackage{booktabs,makecell}

\usepackage{enumitem}
\setlist[enumerate]{nosep, leftmargin=*, label=\textbf{[\arabic*]}, itemsep=2pt}

\title{SCRIBE: Structured Chain Reasoning for Interactive Behavior Explanations using Tool Calling}

\author{
  Fares Fawzi,
  Vinitra Swamy,
  Dominik Glandorf,
  Tanya Nazaretsky,
  Tanja Käser\\
  EPFL \\
  \texttt{\{firstname.lastname\}@epfl.ch} \\
}

\begin{document}
\maketitle
\begin{abstract}

Language models can be used to provide interactive, personalized student feedback in educational settings. However, real-world deployment faces three key challenges: privacy concerns, limited computational resources, and the need for pedagogically valid responses. These constraints require small, open-source models that can run locally and reliably ground their outputs in correct information. We introduce SCRIBE, a framework for multi-hop, tool-augmented reasoning designed to generate valid responses to student questions about feedback reports. SCRIBE combines domain-specific tools with a self-reflective inference pipeline that supports iterative reasoning, tool use, and error recovery. We distil these capabilities into 3B and 8B models via two-stage LoRA fine-tuning on synthetic GPT-4o-generated data. Evaluation with a human-aligned GPT-Judge and a user study with 108 students shows that 8B-SCRIBE models achieve comparable or superior quality to much larger models in key dimensions such as relevance and actionability, while being perceived on par with GPT-4o and Llama-3.3 70B by students. These findings demonstrate the viability of SCRIBE for low-resource, privacy-sensitive educational applications.

\end{abstract}

\section{Introduction}

Education at scale, in contexts like massive open online courses (MOOCs) or large in-person lecture halls, enables thousands of learners to engage with the same material simultaneously \cite{de2015will}. However, this scale comes at a cost: limited access to personalized guidance, feedback, and support. 

Recent progress in Large Language Models (LLMs) offers a promising avenue toward personalized support at scale. LLMs have been applied to a wide range of tasks including question generation \cite{scaria2024automated,2024_Hang, 10.1007/978-3-031-64315-6_25,ma2024teach,DBLP:conf/emnlp/Liang0RC0K23}, 
grading \cite{golchin-etal-2025-grading}, and automatic feedback generation \cite{Phung2023GeneratingHF, 10.1371/journal.pone.0304013, swamy2024explanationsactionzeroshottheorydriven, nair-etal-2024-closing}.

\begin{figure}[t]
  \centering
  \includegraphics[width=1.\linewidth]{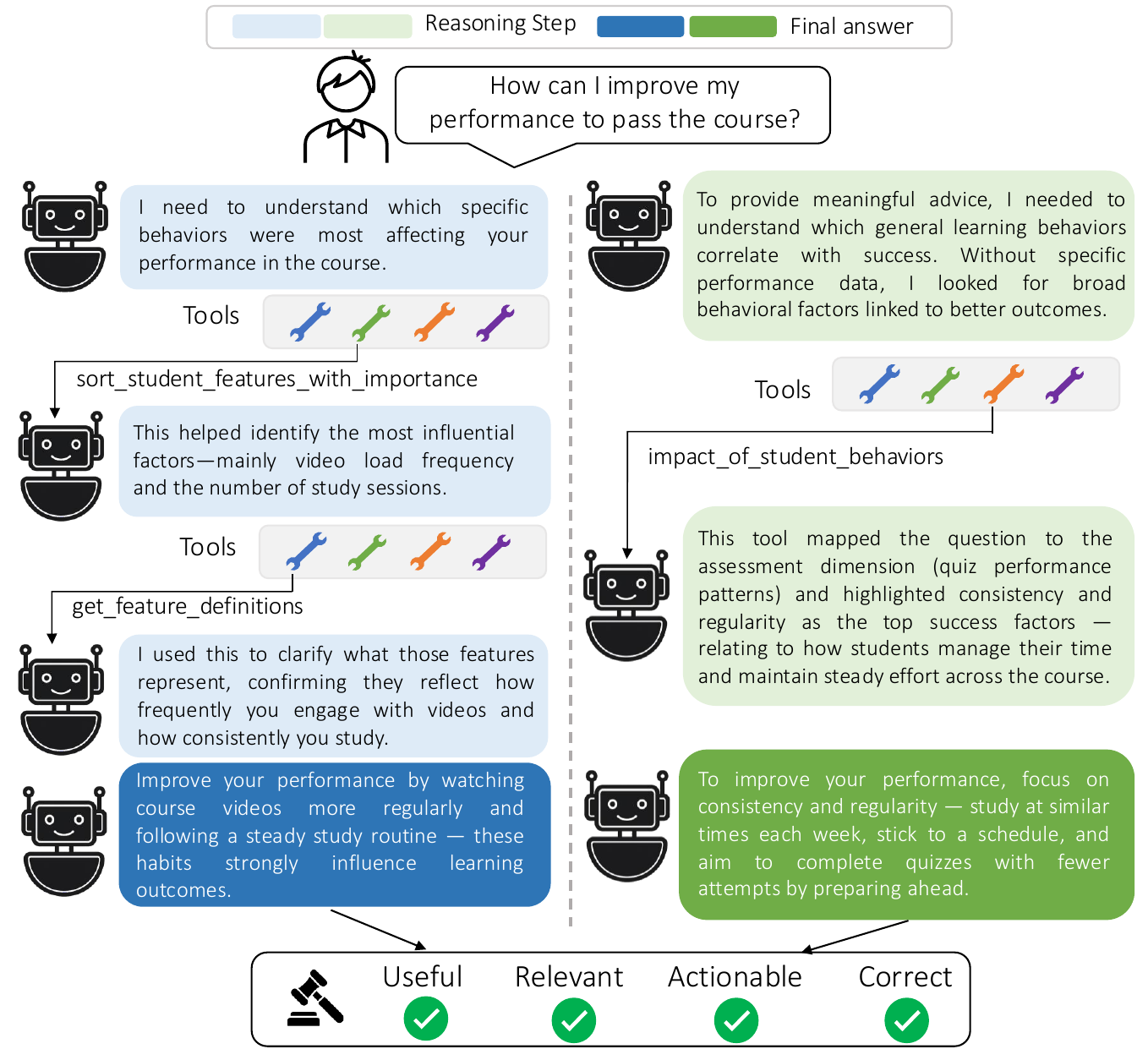}
  \caption{Structured multi-hop reasoning for pedagogically valid feedback via tool calls. The question is addressed using distinct reasoning strategies: one model uses multi-step analysis of learner behavior for a personalized advice (\textit{left}), the other links it to effective learning behavior dimensions for general guidance (\textit{right}).}
   \label{fig:pull_fig}
   \vspace{-6mm}

  \label{fig:overview}
\end{figure}

Despite promising results in educational tasks, LLMs face challenges limiting their reliability in real-world use. Hallucinations and factually incorrect explanations can mislead learners and erode trust, especially problematic in education, where responses must be accurate and pedagogically sound   \cite{10.1007/978-3-031-72315-5_20, manakul-etal-2023-selfcheckgpt, kumar-etal-2023-language, levonian2025designing}. A promising direction to mitigate this is retrieval-augmented generation (RAG) \cite{Fang2025, 10.1145/3686852.3686864}, or tool augmentation \cite{wu2024avatar,ross-etal-2025-when2call, toolformer, gorilla, yao2023react, inaba-etal-2023-multitool} where models use external resources or tools to support reasoning and verification. While these methods improve factuality and interpretability, they are more effective in large models \cite{shen-etal-2024-small} (such as GPT-4o \cite{openai2024gpt4ocard}), which are costly to run. As a result, there is growing interest in training smaller, open-source models that can run locally and securely \cite{zhang-etal-2024-cogenesis}.

Recent work has explored fine-tuning small models on synthetic tool-calling data \cite{gorilla, toolformer, liu2024toolace, qin2023toolllm}. However, these efforts typically address narrow tasks with short, domain-agnostic prompts and a known, fixed sequence of tool calls (e.g., querying the fuel level of an aircraft). This setup fails to reflect real-world domains like education, where open-ended questions require flexible, multi-step reasoning. As shown in Fig.~\ref{fig:pull_fig}, a question like “How can I improve my performance?” can be answered through different tool-use paths. The provided responses are both pedagogically valid, yet created by distinct reasoning trajectories.

In this work, we propose SCRIBE, a framework for self-reflective, multi-hop tool reasoning in educational feedback scenarios, where models must flexibly use external tools and iteratively revise their outputs to generate pedagogically meaningful responses. We collect real student questions about structured feedback reports and augment them with high-quality synthetic data including reasoning traces, tool calls, and final responses. We fine-tune small open-source models via a two-stage LoRA \cite{hu2022lora} pipeline and implement a self-reflective inference loop that enables iterative reasoning and tool use outperforming or matching larger models. Our evaluation combines automatic assessment using a human-aligned GPT-as-a-judge, alongside a user study with 108 students interacting with feedback across three different MOOCs. Notably, we find students equally rate our SCRIBE-trained 8B model, a much larger Llama-3.3 70B and GPT-4o. Our main contributions are:

\begin{enumerate} [label=\arabic*.] 
\item \textbf{We propose SCRIBE, a framework for multi-hop tool reasoning}, where models must flexibly call tools and self-reflect to generate high-quality responses.
\item \textbf{We distill tool calling and self-reflection reasoning behavior of a larger model (GPT-4o) into relatively smaller open-source models} through a two-stage LoRA fine-tuning process to enhance reasoning and multi-hop tool calling.
\item \textbf{We create a new synthetic dataset of 7000 student performance feedback questions} derived from 28 real-world students with answers, tool calling and reasoning chains. 

\item \textbf{We design a rubric for interactive feedback evaluation for a human-aligned GPT-as-a-judge}, enabling scalable and consistent evaluation of model responses.

\item \textbf{We conduct a real-world interactive user study with 108 university students} assessing  perception of interactions with a small SCRIBE 8B model, Llama-3.3 70B, and GPT-4o across distinct reports from three different MOOCs.
\end{enumerate}

\noindent We provide our full implementation, open-source dataset, and trained models, enabling reproducibility and further research.\footnote{All resources are available at \href{https://github.com/epfl-ml4ed/SCRIBE}{https://github.com/epfl-ml4ed/SCRIBE}.}

\begin{figure*}[t]
  \centering
  \includegraphics[width=\linewidth]{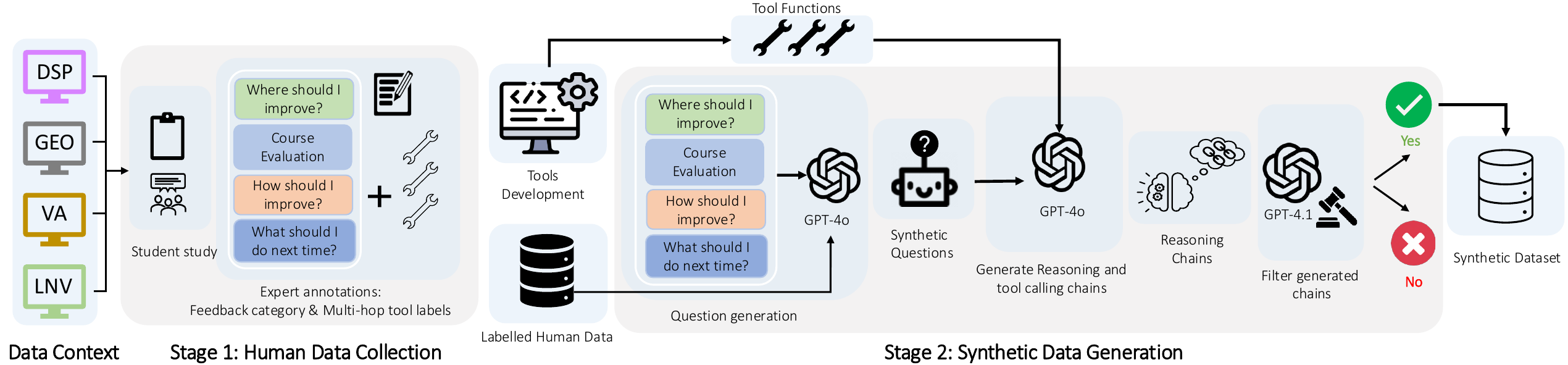}
  \caption{\textbf{SCRIBE Data Generation Pipeline}. Synthetic data is generated by collecting questions from students to guide expert annotators in identifying essential tools (Stage 1). GPT-4o  generates reasoning chains with these tools, and GPT-4.1 filters the outputs based on actionability, relevance, tool use, and correctness (Stage 2).  }
  \vspace{-4mm}
  \label{fig:method_data_fig}
\end{figure*}
\vspace{-1mm}
\section{Related Work}
\vspace{-1mm}
\textbf{Tool-Augmented Language Models.} Tool calling helps LLMs compensate for missing world knowledge and reduce hallucinations \cite{komeili-etal-2022-internet,10.1145/3626772.3661381}. Recent work has explored in-context learning and few-shot prompting to encourage reasoning about tool use \cite{yao2023react, kim2024llm, shen2023hugginggpt, chen-etal-2023-chatcot}. Prompting techniques like chain-of-thought (CoT) \cite{wei2022chain}, and ReAct \cite{yao2023react} structure intermediate reasoning and improve factuality (as demonstrated by \citet{inaba-etal-2023-multitool}), but remain fragile in smaller models and generalize poorly with weak instruction-following.

To enhance tool calling, especially in smaller open-source LLMs, other works have performed finetuning. Toolformer \cite{toolformer} uses a self-supervised approach with LLM-generated data to train models to decide when to call APIs. Gorilla \cite{gorilla} fine-tunes a LLaMA-based model on GPT-4 instruction–API pairs to generate accurate calls from documentation or internal knowledge. Recent works like ToolLLM \cite{qin2023toolllm} and ToolACE \cite{liu2024toolace} use synthetic data to support multi-tool use for complex tasks. However, tool use is often treated as an end in itself rather than a step toward producing high-quality, correct answers. Despite gains in tool call accuracy, models are rarely trained to reason before and after tool calls, and are seldom evaluated in domain-specific, real-world settings such as educational feedback where clarity, correctness, and user trust are essential. As a result, their responses may often lack coherence, context-awareness, and alignment with user needs.

\noindent\textbf{LLMs in Education.}
LLMs are increasingly used in education, enabling natural interactions through conversational agents \cite{lieb2024student, wolfbauer2023rebo, neumann2024llm, 10.1145/3657604.3662041}. Their broad domain knowledge reduces reliance on domain-specific models, supporting applications like personalized learning \cite{park2024empowering}, knowledge tracing \cite{neshaei2024towards}, and automated feedback \cite{stamper2024enhancing}. Prior work has explored various integration strategies, often focusing on \textit{prompting}, e.g., zero-shot prompts for automatic science scoring \cite{wu2023matching} or CoT for classifying learning outcomes via Bloom’s taxonomy \cite{almatrafi2025leveraging}. Others fine-tune LLMs on educational data, e.g., to recognize epistemic and topic-related dialogue acts in collaborative learning \cite{acosta2024recognizing} or to score math responses \cite{Morris2024}. Prior work also explored RAG, using textbooks for guidance \cite{henkel2024retrieval} or student reflections for feedback \cite{neshaei2025user}. However, most models act as \textit{standalone} generators, with few integrating tools for grounded interactions.

\begin{figure*}[t]
  \centering
  \includegraphics[width=\linewidth]{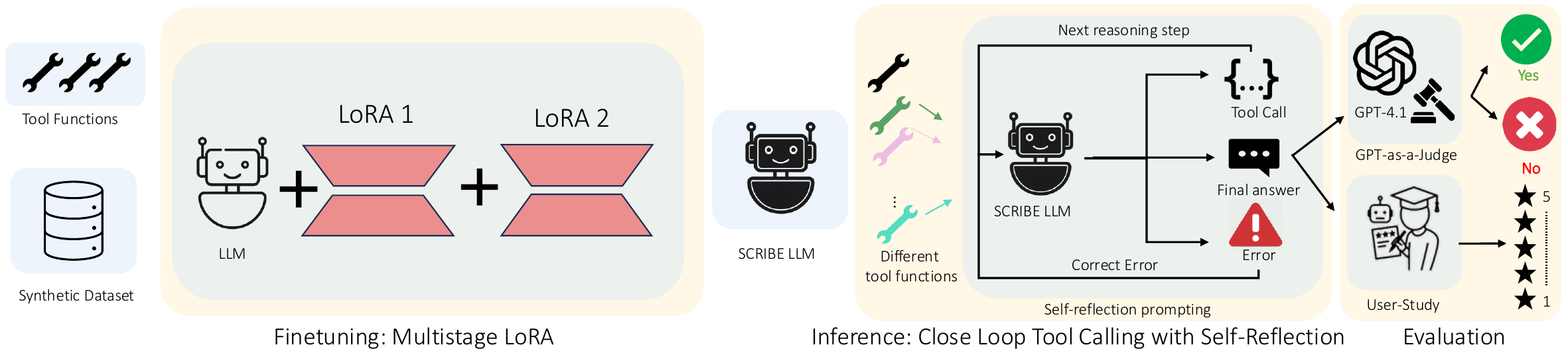}
  \caption{\textbf{SCRIBE finetuning, inference, and evaluation pipelines}. Finetuning involves two successive LoRA stages for multi-hop reasoning with tool use. Inference operates as a closed-loop system with self-reflection prompting for error correction. Evaluation combines GPT-as-a-judge assessments and a user study.}
  \label{fig:method_finetuning_fig}
    \vspace{-3mm}
\end{figure*}
\vspace{-1mm}
\section{Methods}
\vspace{-1mm}
Our goal is to enable interactive feedback with small LLMs by using multi-hop tool calling to generate pedagogically meaningful personalized responses. Our framework, SCRIBE, consists of two main phases: (1) Dataset generation (see \cref{fig:method_data_fig}) and (2) Finetuning and inference (see \cref{fig:method_finetuning_fig}).

\subsection{Dataset Generation Pipeline}\label{sec:dataset}
Our dataset generation pipeline consists of (1) a user study to identify real student questions and categorize them by pedagogical need, (2) domain-specific tools to support grounded, context-aware answers, (3) synthetic data generation using GPT-4o simulating multi-hop reasoning and tool calls. 

\subsubsection{Data Context} \label{sec: problem_setup}
Our experiments use data from four globally-offered MOOCs at a European university: Digital Signal Processing (DSP), Éléments de Géomatique (GEO), Villes Africaines (VA), and Launching New Ventures (LNV). Each includes weekly video lectures, quizzes, and graded assignments. To analyze student performance, we use feedback reports from iLLuMinaTE \cite{swamy2024explanationsactionzeroshottheorydriven}, a zero-shot LLM-XAI framework that generates social science theory-driven, actionable explanations based on behavioral features predicting pass/fail outcomes. We focus on feedback based on social science theories and post-hoc explainers shown to be highly useful and actionable: Necessity and Robustness selection (NR) \cite{Lipton_1990, LOMBROZO2010303}, Abnormal Conditions (AC) \cite{Abnormal_Conditions}, and Contrastive Explanation (Con) \cite{contrastive_explanation}, with Contrastive Explanation Method (CEM) \cite{dhurandhar2018explanations} as the explainer.

\subsubsection{Human Data Collection} \label{sec: human_data_collection}
\textbf{Student Study.} 
To design an interactive feedback system, we first investigated the types of questions students ask when presented with explanation-based feedback. We used five feedback reports from \citet{swamy2024explanationsactionzeroshottheorydriven}. Two reports described a student enrolled in DSP (based on the NR and Con theories), two reports belonged to a student from GEO (again one report per theory), and one report was from a student in VA using the AC theory.

We conducted a study with 28 postgraduate STEM students, each randomly assigned one of five reports and given a brief description of the associated MOOC. Participants (1) wrote three follow-up questions about the feedback, (2) rated five GPT-4o-generated questions on a 1–5 scale ($5=$ very useful), and (3) selected the most useful feedback category, from \citet{hattie}: What have I done well?, Where should I improve?, How should I improve?, and What should I do next time?. All students gave informed consent to participation and the study was approved by the university's human research ethics commission.

\noindent \textbf{Expert Annotations.}
We manually annotated 75 student-written questions categorizing feedback students seek, using 3 dimensions from \citet{hattie}: Where to improve? (\textit{Where}?), How to improve? (\textit{How}?), and What to do next time? (\textit{Next Time}). Our rubric is provided in \cref{appx:anno_rubric}. Two expert annotators independently labeled the questions, achieving substantial agreement (Cohen’s $\kappa = 0.67$). During annotation, we identified an additional category, \textit{"Course Evaluation"}, for questions about course structure and assessment. Based on annotations, we derived six tools needed to meaningfully answer these queries.

\subsubsection{Tools Development}
\label{Methods:Tool Development}
To be able to answer the students' questions, we developed six different domain-specific tools.

\noindent \textbf{Textbook and Syllabus Retrieval Tools.}
For course content questions, we used RAG over MOOC materials. Textbook sections and exercises were embedded using the bge-small model \cite{bge_embedding}, enabling query-based retrieval. Syllabi were embedded with the bilingual-embedding-base model \cite{bilingual-embedding-base} for structure-related queries.\\
\noindent \textbf{Topic Dependency Mapping.}
To clarify topic dependencies, we constructed skill maps that capture prerequisite relationships. For DSP, we adopted the map from \citet{2022.EDM-long-papers.9}. For GEO, the instructor provided a custom map. For VA and LNV, we extracted skills from video transcripts using GPT-4o and then re-prompted it to infer dependencies. The VA map was validated by the instructor. Finally, we implemented a function that, given a MOOC name and week, returns the relevant prerequisite weeks. The full set of maps is available in \cref{appx:tools}.

\noindent \textbf{Grade Calculator.} 
To address performance questions, we designed a function that calculates student total grade from their ID, compares to the passing threshold, and returns the points needed to pass.


\noindent\textbf{Sort Student Features.} 
The tool summarizes student progress using behavioral features from \cite{swamy2024explanationsactionzeroshottheorydriven}, and importance ranked by CEM. For a student and week, it returns 5 most and least important features with raw feature values for context.

\noindent \textbf{Features Description Search.} 
Some student questions focused on unfamiliar terms from feedback reports, derived from features used in student modeling \cite{swamy2024explanationsactionzeroshottheorydriven}. To support these queries and the \textit{Sort Student Features} tool, we developed a function that retrieves feature descriptions. Given a feature name, we use efficient fuzzy string matching for an efficient nearest-neighbor matching and return the corresponding definition.

\noindent \textbf{Student Behavior Impact on Performance.}
The tool answers hypothetical questions about how behavioral changes affect outcomes (e.g., “Would more consistent engagement improve my grade?”). Given a MOOC name and query, it maps the input to one of five behavioral dimensions \cite{10.1007/978-3-031-11644-5_8}—Effort, Consistency, Proactivity, Assessment, and Regularity—linked to features from \cite{swamy2024explanationsactionzeroshottheorydriven} using CEM-derived importance scores. Queries and feature descriptions are embedded with all-MiniLM-L6-v2 \cite{reimers-2019-sentence-bert} and matched via cosine similarity. The tool returns the closest dimension and two alternatives, each with a brief definition, helping students assess their behavior’s impact and explore other strategies.

\subsubsection{Synthetic Data Generation} 
\label{sec: methods-data_gen}
To generate synthetic questions that closely resemble those written by students, we selected 16 students across three MOOCs (DSP, GEO, and VA) and chose two reports per student, each generated using one of two theories introduced in \cref{sec: problem_setup} (NR and Con). For each of these reports, we used real student-written questions and insights collected from students in human study \cref{sec: human_data_collection} to construct the prompts for GPT-4o. We generated 20 synthetic questions per feedback category, per report, yielding a rich dataset of approximately 7000, diverse student-like questions.

Using the generated questions, we prompted GPT-4o with a feedback report and a student question to generate structured reasoning followed by an initial tool call. This tool call is executed, and its output is returned to GPT-4o to produce the next reasoning step and either a subsequent tool call or a final answer. This process is repeated until a final answer is produced. Each example thus forms a reasoning trajectory of alternating reasoning and tool interactions, which we automatically filter using GPT-as-a-judge to exclude samples with erroneous reasoning chains or tool misuse. We used examples that the judge marked YES in all categories (details described in \cref{sec:evaluation}). 

To assess the similarity and diversity of the generated questions relative to real student questions, we first compared the distributions of question lengths and removed outliers that were shorter or longer than student responses. Next, we computed the distributions of Shannon entropy (to estimate token-level information content) and perplexity (to approximate linguistic fluency), and compared these between real and synthetic questions using Jensen-Shannon Divergence (JSD). We performed these comparisons across question types and courses. To further assess semantic diversity, we computed pairwise cosine similarity within each dataset (real and synthetic) across all questions, for each course and feedback category. This enabled us to quantify question diversity within each dataset. \textcolor{black}{Next, to evaluate the similarity of the generated questions to real student questions, we compared the embeddings of 76 generated questions (matched to the number of human-authored ones), using the bge-large-en model \cite{bge_embedding}, against embeddings of (a) real student questions from the same reports and (b) randomly selected SQuAD questions \cite{rajpurkar-etal-2018-know}. We applied Hotelling’s T² test on 2D representations from t-SNE to compare distributions.}

\vspace{-2mm}
\subsection{Inference and Finetuning Pipeline}\label{sec:finetune_inference}
The objective of this pipeline is to distill GPT-4o tool calling and reasoning capabilities into smaller LLMs through a two-stage LoRA finetuning. Our finetuning and inference pipeline consists of 
(1) a multi-stage fine-tuning process where relatively small open-source models (e.g., Llama 8B) are trained via LoRA adapters to perform structured reasoning and tool use, and (2) a closed-loop inference pipeline that supports iterative tool use, self-reflection, and error correction.

\subsubsection{Multi-Stage LoRA Fine-Tuning} \label{sec:Multi-Stage LoRA}
To enhance the reasoning and multi-hop tool use abilities of relatively small open-source models, we distill structured tool-calling behavior from a much larger teacher model (GPT-4o). Inspired by multi-stage instruction tuning and curriculum-style learning \cite{chen2023curriculum, guan-etal-2025-multi, pang-etal-2024-phased}, our training process is divided into two sequential stages that progressively increase task complexity. Each training instance consists of a student query $q$, a feedback report $f$, a sequence of reasoning steps $\{r_i\}_{i=0}^n$, tool calls $\{t_i\}_{i=0}^n$, tool outputs $\{o_i\}_{i=0}^n$, and a final answer $a$. 

\noindent \textbf{Stage 1 (Initial Reasoning and Tool Selection).}
The model is trained to generate an initial reasoning step $r_0$ and the first tool call $t_0$ conditioned on $(q, f)$. This teaches the model how to interpret student questions and initiate tool-call reasoning.

\vspace{-2mm}
\begin{equation}
    r_0, t_0 \sim P_{\text{stage1}}(r, t \mid q, f)
\end{equation}

\vspace{1pt} \noindent \textbf{Stage 2 (Multi-Hop Reasoning and Answer Generation).}
Conditioned on $q$, $f$, the initial tool call $t_0$ and output $o_0$, the model learns to iteratively reason and revise its outputs across multiple steps. It produces intermediate reasoning steps $r_i$, additional tool calls $t_i$, and the final answer $a$.

\vspace{-4mm}
\begin{align}
r_i, t_i &\sim P_{\text{reason}}\left(r, t \mid q, f, \{(r_j, t_j, o_j)\}_{j < i}\right), \notag \\
&\quad \text{for } i = 1, \ldots, n \\
a &\sim P_{\text{answer}}\left(a \mid q, f, \{(r_j, t_j, o_j)\}_{j \leq n}\right) \notag
\end{align}

This decomposition ensures the model first learns how to initiate tool-augmented reasoning before handling more complex reasoning trajectories with iterative refinement. We use LoRA adapters for efficient parameter updates in both stages.

\subsubsection{Closed-Loop Tool Calling}
Inspired by AnyTool \cite{10.5555/3692070.3692540} which re-queries the tool using a self-reflection loop, we implement self-reflective, multi-hop reasoning as our \textbf{prompting framework for inference}, where the model incrementally constructs responses to student questions by interacting with external tools and revising reasoning based on their outputs. We provide the prompts in \cref{appx:self-reflection-prompt}. This task is inherently underdetermined, as different sequences of tool calls may lead to equally valid answers. Our pipeline supports this flexibility while enabling error recovery and iterative refinement.

Formally, for a given student query $q$ and feedback report $f$, the model produces an initial reasoning step $r_0$ and a corresponding tool call $t_0$. The output $o_0$ from executing $t_0$ is passed back to the model, which generates the next reasoning step $r_1 = \texttt{Reason}(r_0, o_0, q, f)$, followed optionally by another tool call $t_1$. This process continues for up to $N$ steps, producing a trajectory:
\vspace{-1mm}
\begin{equation}
    (f, q, r_0, t_0, o_0, r_1, t_1, o_1, \dots, r_n, a)
\end{equation}
where $a$ is the final answer and $n < N$. At each step $i$, the model decides whether to call another tool or produce a final answer, based on the evolving context of the query, feedback report, previous reasoning steps, and tool outputs. This iterative process continues until the model outputs a final answer or reaches a predefined step limit $N$.





The model may select the same tool repeatedly or switch tools across steps, depending on the evolving context. To improve robustness, the system monitors for tool-call errors or instruction violations (e.g., invalid tools, skipped reasoning). In such cases, the model is re-prompted to self-reflect and revise its reasoning or tool choice. If no valid answer is generated after $N$ iterations, the interaction is terminated and marked as unresolved.

\subsubsection{Evaluation}
\label{sec:evaluation}
We evaluated the models' responses using expert annotation and a LLM-as-a-judge protocol as well as through a user study with real students.

\noindent \textbf{GPT-as-a-Judge.} Given the open-ended task, standard metrics like tool selection accuracy are insufficient, as multiple tool sequences can yield valid answers. We therefore developed a rubric to evaluate both the tool used and the model’s student-facing final response. Based on existing literature, we defined four criteria and added a fifth, tool relevance, specific to our setting. The criteria include: (1) \textbf{Relevance} to the question \cite{LLM-as-a-judge}, (2) \textbf{Actionability} in terms of providing concrete advice \cite{swamy2024explanationsactionzeroshottheorydriven}, (3) \textbf{Tool Relevance} (whether the selected tools were appropriate), (4) \textbf{Spelling and Grammar}  \cite{swamy2024explanationsactionzeroshottheorydriven}, and (5) \textbf{Correctness} based on factual alignment with tool outputs and feedback \cite{LLM-as-a-judge}. The detailed rubric is  provided in \cref{appx:GPT-as-a-Judge}.

In a first step, two researchers independently labeled 60 instances comprising 20 responses, tool calls, and tool outputs from three different models (Llama-3.1 8B base, SCRIBE, and Llama-3.3 70B) sampled across three MOOCs (DSP, GEO, and VA). The annotations achieved an overall Cohen’s $\kappa$ of 0.85, indicating strong inter-rater agreement. To assess the quality of model outputs at scale, we then adopted GPT-4.1 \cite{openai2025gpt41} as an third evaluator, following prior work on LLM-based judgment for response quality \cite{liu-etal-2023-g, LLM-as-a-judge,qin2023toolllm, 10.5555/3692070.3692540}. Each judgment is generated by prompting GPT-4.1 with a feedback report, student question, a description of available tools, the model’s full reasoning trace (with tool calls and outputs), and definitions for each evaluation criterion. We used CoT prompting to encourage step-by-step reasoning before GPT-4.1 returns a binary rating (Yes/No) for each question criterion \cite{qin2024infobench}. To ensure reliabilty, we ran GPT-4.1 five times, achieving Cohen’s $\kappa = 0.818 \pm{0.014}$ between the GPT-4.1 judge and the humans. We provide prompts and per criterion inter-annotator agreement in \cref{appx:GPT-as-a-Judge}.

\noindent \textbf{User Study.}
To evaluate how students perceive model-generated responses, we conducted a user study comparing a small multi-stage LoRA-tuned model (ToolACE-8B SCRIBE) to two larger LLMs (Llama-3.3 70B and GPT-4o). To reflect deployment constraints where hosting large models may be infeasible for schools, we used API for Llama-3.3 70B and GPT-4o. We recruited 108 students via Prolific\footnote{\url{https://www.prolific.com}} (see \cref{appx:user-study} for more details). All participants provided informed consent, and the study was approved by our university’s human research ethics commission. Each participant saw three feedback reports (passing and failing students) generated by iLLuMinaTE~\cite{swamy2024explanationsactionzeroshottheorydriven}, each from a different MOOC: DSP, GEO, and LNV (hold-out MOOC). The study was designed to ensure that each participant interacted with reports from all three MOOCs and models. We constructed 108 unique combinations, each consisting of one student report per course (drawn from six possible reports per course: 3 passing and 3 failing), with each report paired with a different model. Report–model assignments were permuted to ensure that each model was used exactly once within each combination and to prevent ordering effects.

Participants posed 3–5 unrestricted questions per report to have natural conversations.
After each conversation, participants rate the model’s responses on a 5-point scale (1 is lowest and 5 is highest) across five criteria from prior work \cite{swamy2024explanationsactionzeroshottheorydriven, 10.1145/3636555.3636898}:
(1) \textbf{Relevance}: Response directly addresses the question. 
(2) \textbf{Usefulness}: Response provides meaningful insights that answer the question and that can enhance learning or deepen understanding.  
(3) \textbf{Actionability}: Response provides clear steps or instructions.
(4) \textbf{Coverage}: Response comprehensively addresses all components of questions asked, including sub-questions.  
(5) \textbf{Conciseness}: Response is clear, and complete with minimal redundancy.

At the end of the study, participants reviewed the three full conversations side by side and selected their overall preferred interaction and provided the reasons for their preference in an open text field.

\noindent \textbf{Generalisation to Unseen Tools.} \textcolor{black}{To assess whether SCRIBE can extend its tool-use behaviour beyond those seen in training, we introduced a new tool, \texttt{web\_search}, designed to retrieve online resources. We evaluated generalisation in two ways. First, we used 27 GPT-4o–generated questions specifically constructed to test whether the model could invoke the unseen \texttt{web\_search} tool after training on a different set of tools. Second, we used the same 192-question test set employed for model evaluations, spanning the three MOOCs (DSP, GEO, VA), and augmented the existing tool set with \texttt{web\_search} as an extra unseen tool. We then compared our ToolACE-8B-SCRIBE with the base ToolACE-8B in a zero-shot setting.}


\vspace{-1mm}
\section{Results}\label{sec: experiments}
\vspace{-1mm}
We conducted a series of experiments to evaluate the quality of the synthetic data used to train SCRIBE, the response quality of the model through a quantitative analysis, and student perception of its outputs through a user study. 

\noindent \textbf{Experimental Protocol.} We finetuned and evaluated three small models: Llama-3.2 3B and Llama-3.1 8B, which natively support tool calling, and ToolACE-8B \cite{liu2024toolace}, an 8B model that achieves state-of-the-art performance on the Berkeley Function Calling Leaderboard (BFCL)  \cite{berkeley-function-calling-leaderboard}, and was able to follow our self-reflection and reasoning instructions. The finetuning required six A100 GPU hours per stage. We compared the small models to GPT-4o (gold standard) and Llama-3.3 70B. All small models were finetuned on 7,000 generated questions (see \cref{sec: methods-data_gen}) with corresponding tool-use and reasoning chains (see \cref{sec:Multi-Stage LoRA}). Our self-reflection inference pipeline was applied uniformly across models for fair comparison. Evaluation was conducted on 192 test questions, including 75 written by real students and additional synthetic questions (unseen in fine-tuning) used to balance coverage across three MOOCs (DSP, GEO, VA) and four categories (How, Where, Next Time, Course Evaluation). We also evaluated on 192 additional held-out questions from the LNV MOOC which was not included in the fine-tuning. For the Llama-3.1 8B and ToolACE-8B models, we achieved best results with LoRAs of rank of 256 (see ablations \cref{appx:ablations}). We used LoRAs of rank of 128 for Llama-3.2 3B.

\vspace{-3mm}
\subsection{Synthetic student questions closely match real student questions} \label{sec: exp_data_gen}
\vspace{-1mm}
To evaluate the quality and variety of GPT-4o-generated questions, we compared them to real student-written questions. Table~\ref{tab:jsd-summary} shows the JSD for the Shannon entropy and for perplexity between student and generated questions as well as cosine similarity within each dataset. We observe that all JSD values are $<0.387$, indicating that the generated questions are reasonably close to human questions in both entropy and perplexity. Among the MOOCs, the lowest divergence in entropy was observed in GEO (entropy JSD = 0.114 ± 0.076), while the highest was in VA (entropy JSD = 0.335 ± 0.144), suggesting more distinctive phrasing in student-written questions for that course. For perplexity, VA had the lowest divergence (0.140 ± 0.093), indicating strong alignment in fluency. Across question categories, \textit{“Next Time”} questions diverged the most (entropy JSD = 0.387 ± 0.089 and perplexity JSD = 0.211 ± 0.064), likely due to the high variability and learner-specific nature of next-step feedback questions \cite{hattie}. The pairwise cosine similarity was slightly higher among generated questions in GEO and DSP and categories \textit{How?} and \textit{Where?}, indicating slightly less variation. However, overlapping standard deviations suggest that both generated and human questions exhibit comparable diversity.

\textcolor{black}{Complementing these distributional metrics, Table~\ref{tab:synthetic_vs_human_single} reports Hotelling’s $T^2$ test results on t-SNE embeddings. 
Generated and human questions are not significantly different ($p=0.229$), whereas both differ significantly from random questions ($p \approx 10^{-16}$). Their centroids also cluster closely in t-SNE space, further confirming that GPT-4o-generated questions align with real student questions while remaining distinct from unrelated out-of-domain data.}

\begin{table}[t]
\centering
\caption{Jensen-Shannon Divergence (JSD) and pairwise cosine similarity between human and generated questions across MOOCs and question categories.}
\label{tab:jsd-summary}
\resizebox{\columnwidth}{!}{
    \begin{tabular}{lccccc}
    \toprule
    \textbf{Group} & \textbf{Type} & \textbf{ \begin{tabular}[c]{@{}c@{}}JSD \\ (Entropy) \end{tabular}  } & \textbf{\begin{tabular}[c]{@{}c@{}}JSD \\ (Perplexity)  \end{tabular}} & \multicolumn{2}{c}{\textbf{\begin{tabular}[c]{@{}c@{}}Pairwise \\ Cosine  Similarity \end{tabular}} } \\
    &&&&Generated&Human\\
    \midrule
    \multirow{3}{*}{MOOC} 
      & GEO & 0.114 $\pm{0.076}$ & 0.202 $\pm{0.079}$ & 0.265 $\pm{0.034}$ & 0.238 $\pm{0.024}$ \\
      & DSP & 0.327 $\pm{0.079}$ & 0.212 $\pm{0.095}$ & 0.279 $\pm{0.044 }$ & 0.265 $\pm{0.029}$\\
      & VA & 0.335 $\pm{0.144}$ & 0.140 $\pm{0.093}$ & 0.280 $\pm{0.064}$ & 0.307  $\pm{ 0.047}$ \\  
    \midrule
    \multirow{3}{*}{\begin{tabular}[c]{@{}c@{}}Question \\ Category \end{tabular}}
      & How? & 0.180 $\pm{0.093}$ & 0.184 $\pm{0.095}$ & 0.241 $\pm{0.035}$ & 0.234 $\pm{0.034}$\\
      & Where? & 0.242 $\pm{0.121}$  & 0.152 $\pm{0.075}$ & 0.272 $\pm{0.046}$ & 0.249 $\pm{0.021}$ \\
      & Next Time & 0.387 $\pm{0.089}$ & 0.211 $\pm{0.064}$ &  0.271 $\pm{0.052}$ & 0.319 $\pm{0.026}$ \\
    \bottomrule
    \end{tabular}}
\end{table}

\begin{table}[t]
\centering
{\color{black}
\caption{ 
\textcolor{black}{Synthetic vs.\ human question similarity. Left: descriptive statistics of t-SNE (2D) embeddings per questions source (Generated, Human, Random). Right: Hotelling’s $T^2$ tests with $F$ and $p$ values for pairwise comparisons.}
}
\label{tab:synthetic_vs_human_single}
\resizebox{\columnwidth}{!}{
\begin{tabular}{lccc ccccc}
\toprule
\multicolumn{4}{c}{\textbf{Descriptive (t-SNE 2D)}} & & \multicolumn{4}{c}{\textbf{Hotelling’s $T^2$ Tests}} \\
\cmidrule(lr){1-4} \cmidrule(l){6-9}
\textbf{Metric} & \textbf{Generated} & \textbf{Human} & \textbf{Random} & & \textbf{Pair} & $\mathbf{T^2}$ & $\mathbf{F}$ & $\mathbf{p}$ \\
\midrule
Centroid $(x,y)$
& [$-2.19$, $\,9.85$]
& [$-0.32$, $\,8.88$]
& [$3.22$, $-15.03$]
& &
Gen vs.\ Human   & 2.99   & 1.49   & 0.229 \\
STD $(x,y)$
& [$7.64$, $\,8.85$]
& [$7.78$, $\,6.96$]
& [$8.62$, $\,8.64$]
& &
Gen vs.\ Random  & 484.62 & 241.28 & $1.11 \times 10^{-16}$ \\
SEM $(x,y)$
& [$0.71$, $\,0.82$]
& [$0.89$, $\,0.80$]
& [$0.79$, $\,0.79$]
& &
Human vs.\ Random & 440.34 & 219.04 & $1.11 \times 10^{-16}$ \\
\bottomrule
\end{tabular}}
}
\end{table}




\begin{tcolorbox}[
  colback=gray!10, 
  colframe=gray!50, 
  boxrule=0.5pt, 
  arc=2pt,
  fontupper=\small
]
GPT-4o-generated questions closely match real student ones in fluency, content, and diversity, validating them as high-quality training data.
\end{tcolorbox}

\subsection{SCRIBE achieves the performance of significantly larger models} \label{sec: main_results}

\begin{figure}[t]
  \centering
  \includegraphics[width=\linewidth]{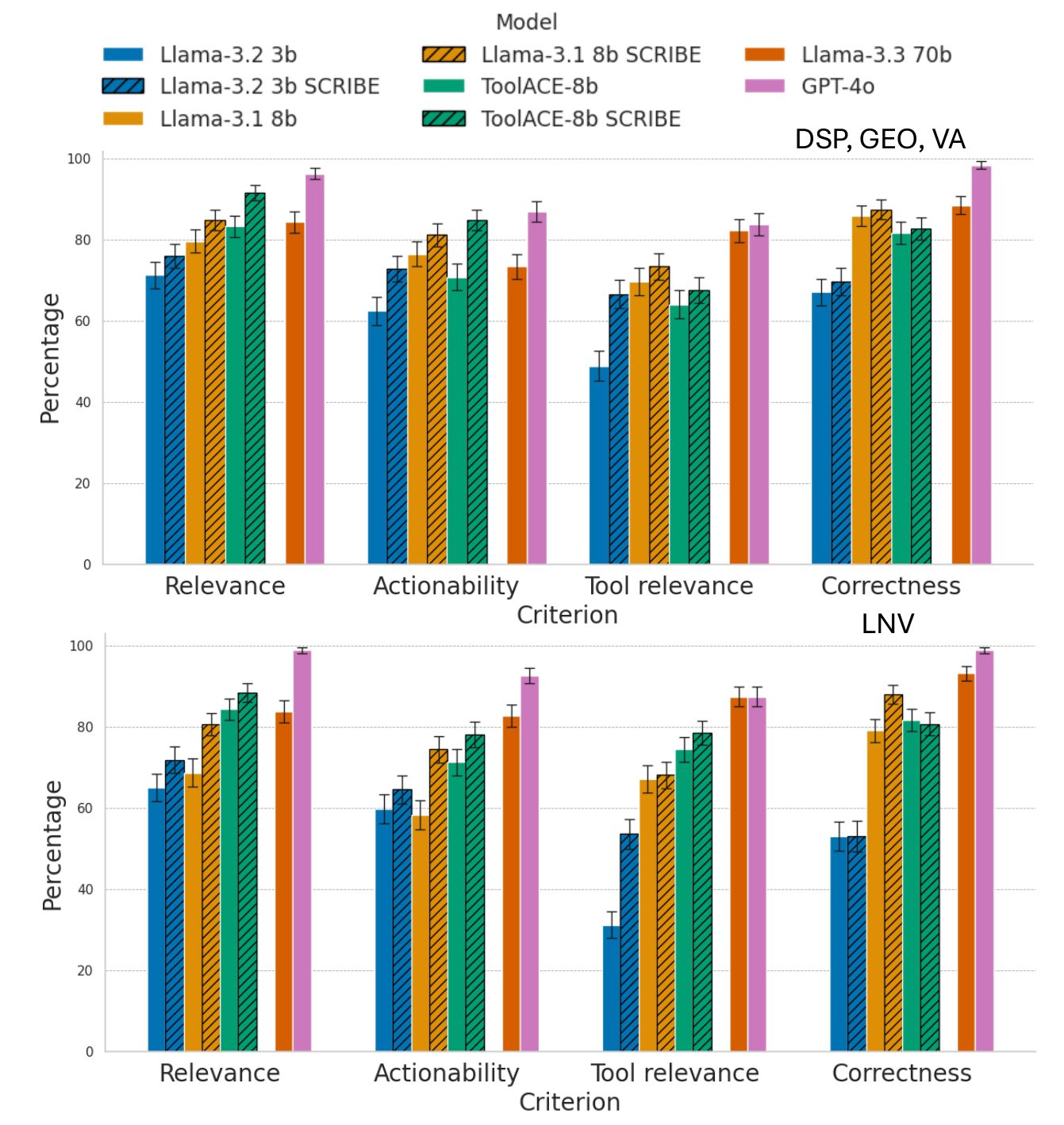}
  \caption{Percentage of YES given by GPT-Judge for each criterion on a holdout dataset of GEO, DSP and VA MOOCs (top) and a holdout set of LNV MOOC (bottom). Hashed bars indicate SCRIBE models}
  \label{fig:judge_results}
  \vspace{-3 mm}
\end{figure}

The top plot in Fig.~\ref{fig:judge_results} shows evaluation results on the test dataset from GEO, DSP, and VA. 
Across these courses, fine-tuned SCRIBE models significantly outperform their base versions on relevance, actionability, and tool relevance, with no significant difference in correctness (see Table ~\ref{tab:fisher_seen} in appendix \ref{appx: full_results} for full test results). ToolACE-8B-SCRIBE and Llama-3.1 8B-SCRIBE both surpass the much larger Llama-3.3 70B in actionability, and match it on relevance and correctness. 
The 70B model remains significantly stronger on tool relevance, while improvements in correctness remain modest overall.

The bottom plot of Fig.~\ref{fig:judge_results} shows results on LNV (an unseen MOOC), where a similar pattern holds. 
SCRIBE models again significantly outperform their base counterparts on all criteria except correctness. 
Relative to the 70B model, 8B SCRIBE models show no significant difference in relevance or actionability, but the 70B remains stronger in tool relevance and correctness. 
These findings confirm that SCRIBE finetuning yields statistically robust gains over base models and narrows the gap to much larger systems. Detailed statistical test results are provided in Table~\ref{tab:fisher_lnv} in appendix \ref{appx: full_results}. All models achieved a perfect score on the spelling and grammar criterion, we therefore omitted this category in the Figures.

\begin{tcolorbox}[
  colback=gray!10, 
  colframe=gray!50, 
  boxrule=0.5pt, 
  arc=2pt,
  fontupper=\small
]
SCRIBE-trained models significantly outperform their base versions on relevance, actionability, and tool relevance, while matching much larger models in relevance and actionability.
\end{tcolorbox}

\subsection{Students rate SCRIBE responses highly} 
\label{sec: user_study}
\begin{figure}[ht]
  \centering
  \includegraphics[width=\linewidth]{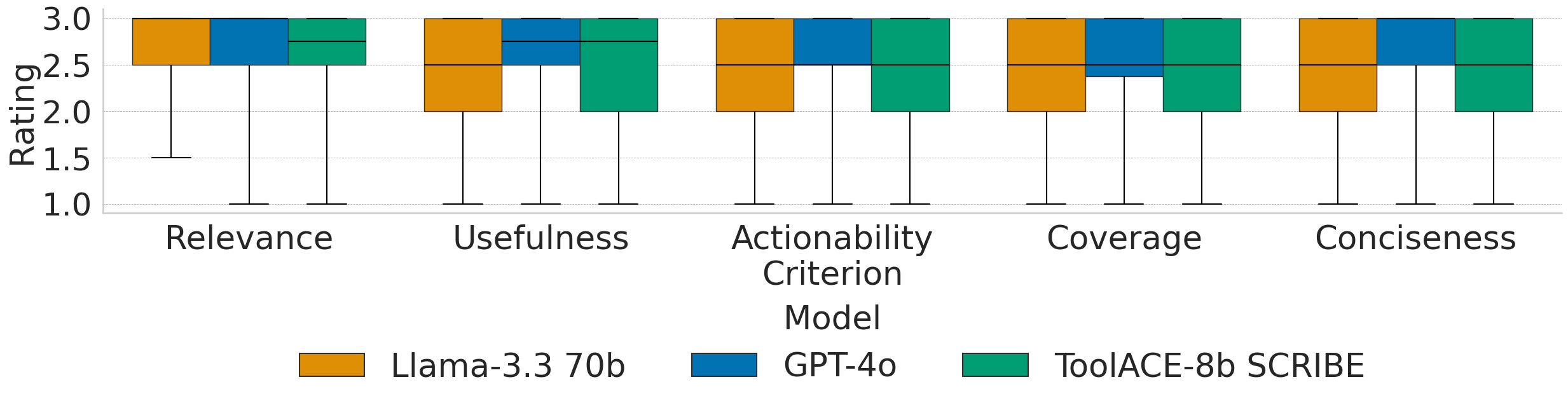}
  \caption{Average ratings from 108 students (1–5 scale) for LLama-3.3 70B, GPT-4o and ToolACE-8B SCRIBE.}
  \label{fig:user_study_rating}
  \vspace{-4 mm}
\end{figure}

 Fig.~\ref{fig:user_study_rating} shows the average ratings per criterion for each model included in the user study. We observe that the ratings across all five criteria are highly similar across models. Despite the SCRIBE model being significantly smaller in size (8B vs. 70B), students perceive its response quality as on par with much larger models. To test whether any observed differences in ratings were statistically significant, we conducted a one-way ANOVA for each criterion across the three models. In all cases, we failed to reject the null hypothesis ($p > 0.05$), indicating no significant difference in perceived response quality (see appendix \ref{appx:anova} for ANOVA results).

When students were asked to select their preferred conversation and explain why, 47.2\% chose GPT-4o, while the remaining responses were evenly split between Llama-3.3 70B and ToolACE-SCRIBE. Among those who preferred GPT-4o, about 25\% cited its detailed explanations as the main reason. Others highlighted its actionable advice and clarity. In contrast, 32.1\% of students who preferred ToolACE-SCRIBE praised its conciseness. One participant stated: \textit{“The feedback provided clear and direct answers to all my questions in a precise and concise manner, making it easy to understand what I’m doing well.”}.

\begin{tcolorbox}[
  colback=gray!10, 
  colframe=gray!50, 
  boxrule=0.5pt, 
  arc=2pt,
  fontupper=\small  
]

Students rate Relevance, Usefulness, Actionability, Coverage, and Conciseness of the \textbf{SCRIBE} model on par with larger API-based models, validating its use in low-resource, privacy-sensitive educational settings.



\end{tcolorbox}

\subsection{SCRIBE generalises to unseen tools}
\textcolor{black}{
On 27 GPT-4o–generated questions specifically designed to trigger the unseen \texttt{web\_search} tool, ToolACE-8B SCRIBE invoked it 9 times, showing that the model can generalise tool-use behaviour in a zero-shot setting. On the original 192-question dataset (DSP, GEO, VA) with \texttt{web\_search} available, ToolACE-8B-SCRIBE used the tool 25 times compared to 7 times for the ToolACE-8B, indicating generalization of tool-use behaviour.
}

\textcolor{black}{
As shown in Table~\ref{tab:toolace_comparison}, both models employed the new tool despite no prior exposure, with SCRIBE achieving higher Tool Relevance. 
However, introducing \texttt{web\_search} led to a slight drop in performance across metrics for both models relative to their runs without it. This is likely due to the larger action space and added ambiguity, since most questions in the 192-question set did not require this new tool.
}\textcolor{black}{
We also examined how the original tools were used and found that each was invoked at least once, underscoring that all were necessary to address student questions. Full results are reported in Appendix~\ref{appx:tool_contributions}.
}

\begin{table}[htbp]
\centering
{\color{black}
\normalsize
\resizebox{\columnwidth}{!}{%
\begin{tabular}{llcccc}
\toprule
\textbf{Group} & \textbf{Type} & \textbf{Relevance} & \textbf{Actionability} & \textbf{Tool Relevance} & \textbf{Correctness} \\
\midrule
\multirow{3}{*}{Web Search} 
& ToolACE-8B SCRIBE-27-Trigger-Qs   & 85.19 $\pm$ 6.95 & 92.59 $\pm$ 5.07 & 77.78 $\pm$ 7.60 & 77.78 $\pm$ 7.70 \\
& ToolACE-8B-Original-192-Qs         & 81.25 $\pm$ 2.80 & 73.44 $\pm$ 3.20 & 64.58 $\pm$ 3.39 & 76.04 $\pm$ 2.90 \\
& ToolACE-8B SCRIBE-Original-192-Qs  & 83.33 $\pm$ 2.71 & 72.40 $\pm$ 3.27 & 73.96 $\pm$ 3.24 & 76.56 $\pm$ 2.95 \\
\midrule
\multirow{2}{*}{No Web Search} 
& ToolACE-8B-Original-192-Qs        & 83.33 $\pm$ 2.80 & 70.83 $\pm$ 3.31 & 64.06 $\pm$ 3.52 & 81.77 $\pm$ 2.81 \\
& ToolACE-8B SCRIBE-Original-192-Qs  & 91.67 $\pm$ 1.88 & 84.90 $\pm$ 2.57 & 67.71 $\pm$ 3.45 & 82.81 $\pm$ 2.76 \\
\bottomrule
\end{tabular}%
}
}
\caption{
\textcolor{black}{GPT-as-Judge evaluation on the 27 new questions that are designed to trigger the (web\_search) tool (27-Trigger-Qs), and on the original with DSP, GEO, and VA (Original-192-Qs) after introducing the new tool}
}
\label{tab:toolace_comparison}
\end{table}

\begin{tcolorbox}[
  colback=gray!10, 
  colframe=gray!50, 
  boxrule=0.5pt, 
  arc=2pt,
  fontupper=\small
]
 SCRIBE demonstrates zero-shot generalisation by successfully invoking \texttt{web\_search}, a tool not seen during training.
\end{tcolorbox}
\vspace{-1mm}
\section{Conclusion}
\vspace{-1mm}
We introduce \textbf{SCRIBE}, a framework for interactive student behavior explanations that combines synthetic data generation, two-stage LoRA fine-tuning, and automatic evaluation with a human-aligned GPT-as-a-Judge. SCRIBE enables small language models to perform self-reflective, multi-hop tool-calling in domains with multiple valid tool-use paths. In education, SCRIBE-trained models consistently outperform base models in relevance, actionability, and tool relevance, while 8B-SCRIBE models match or exceed much larger ones in relevance and actionability, key dimensions of student-centered feedback. A user study with 108 students confirmed they are perceived as equally helpful, relevant, and actionable as larger models. These results show that synthetic data and staged fine-tuning can distill complex tool use into smaller, privacy-preserving educational assistants. Future work will extend SCRIBE to additional models and contexts, and focus on improving correctness and tool relevance.
One possible context is medical and psychiatric diagnosis where different diagnostic paths are valid and lead to the same  diagnosis~\cite{ALARCÓN_2009,Maung_2016,book_diagnosis}. 

\section{Limitations}
While SCRIBE advances small LLMs on interactive student feedback, multihop reasoning, and tool-calling, there is room for further improvement. 
Specifically, gains in correctness remain limited due to the already strong performance of the base models, and tool relevance is another challenging criterion since it depends heavily on the model’s initial reasoning. 
\textcolor{black}{Moreover, while our user study found no perceived difference in the quality of responses between SCRIBE models and much larger API-based models such as Llama-3.3 70B and GPT-4o, we did not evaluate the impact of these models on actual educational outcomes. Assessing how interaction with our system influences student learning and performance remains an important direction for future work.}
\section{Acknowledgments}
We acknowledge that the use of AI assistants (ChatGPT) was limited to polishing the language of the original paper. It was used solely for proofreading and refining grammar, spelling, and phrasing.
\bibliographystyle{acl_natbib}
\bibliography{custom}
\appendix

\section{Statistical Analysis of GPT-as-Judge Evaluation Results} \label{appx: full_results}
To complement the main results reported in Section~\ref{sec: main_results}, we provide the outcomes of statistical significance testing. 
We used Fisher’s Exact Tests to compare (i) SCRIBE models against their corresponding base models, and (ii) 8B SCRIBE models against the Llama-3.3 70B model. 
These tests were conducted on both the original evaluation dataset drawn from DSP, GEO, and VA, and on the unseen LNV dataset. 

Table~\ref{tab:fisher_seen} presents results for DSP, GEO, and VA, showing that SCRIBE models significantly outperform their base versions in relevance, actionability, and tool relevance, with no significant difference in correctness. 
In comparison with the 70B model, 8B SCRIBE achieves significantly higher actionability, parity on relevance and correctness, and lower tool relevance. Table~\ref{tab:fisher_lnv} reports results for the unseen LNV course. 
Here, SCRIBE models again significantly outperform their base versions in relevance, actionability, and tool relevance, while correctness shows no significant difference. 
Against the 70B model, however, the 8B SCRIBE models show no significant difference in relevance and actionability but are significantly weaker in tool relevance and correctness. 

\begin{table}[htbp]
\centering

\renewcommand{\arraystretch}{2} 
\fontsize{50}{35}\selectfont
\resizebox{\columnwidth}{!}{%
\begin{tabular}{lcccc}
\toprule
\textbf{Criterion} &
\textbf{\begin{tabular}[c]{@{}c@{}}SCRIBE vs. Base Models \\ (Odds Ratio, p-value)\end{tabular}} &
\textbf{Interpretation} &
\textbf{\begin{tabular}[c]{@{}c@{}}8B SCRIBE vs. 70B \\ (Odds Ratio, p-value)\end{tabular}} &
\textbf{Interpretation} \\
\midrule
Relevance      & 1.492 (0.0103) & Significantly higher odds for SCRIBE & 1.395 (0.1917) & No significant difference \\
Actionability  & 1.684 (0.00018) & Significantly higher odds for SCRIBE & 1.775 (0.0081) & 8B SCRIBE significantly better \\
Tool Relevance & 1.445 (0.00364) & Significantly higher odds for SCRIBE & 0.516 (0.0023) & 70B significantly better \\
Correctness    & 1.111 (0.5139)  & No significant difference            & 0.742 (0.3048) & No significant difference \\
\bottomrule
\end{tabular}%
}
\caption{Fisher’s Exact Test results on DSP, GEO, and VA (192 questions). Odds ratios $>$ 1 favor the first model listed. Statistically significant differences ($p < 0.05$) are reflected in the interpretation.}
\label{tab:fisher_seen}
\end{table}

\begin{table}[htbp]
\centering

\renewcommand{\arraystretch}{2} 
\fontsize{50}{30}\selectfont
\resizebox{\columnwidth}{!}{%
\begin{tabular}{lcccc}
\toprule
\textbf{Criterion} &
\textbf{\begin{tabular}[c]{@{}c@{}}SCRIBE vs. Base Models \\ (Odds Ratio, p-value)\end{tabular}} &
\textbf{Interpretation} &
\textbf{\begin{tabular}[c]{@{}c@{}}8B SCRIBE vs. 70B \\ (Odds Ratio, p-value)\end{tabular}} &
\textbf{Interpretation} \\
\midrule
Relevance      & 1.54 (0.0027) & SCRIBE significantly better & 1.06 (0.81)  & Not significant \\
Actionability  & 1.53 (0.0010) & SCRIBE significantly better & 0.67 (0.09)  & Not significant \\
Tool Relevance & 1.48 (0.0016) & SCRIBE significantly better & 0.40 (0.0001) & 70B significantly better \\
Correctness    & 1.14 (0.35)   & Not significant             & 0.39 (0.0022) & 70B significantly better \\
\bottomrule
\end{tabular}%
}
\caption{Fisher’s Exact Test results on the unseen MOOC (LNV, 192 questions). Odds ratios $>$ 1 favor the first model listed. Statistically significant differences ($p < 0.05$) are reflected in the interpretation.}
\label{tab:fisher_lnv}
\end{table}
\section{Ablation Studies} \label{appx:ablations}
It is worth noting that in all of our quantitative results we found that Spelling and Grammar was always perfect across all models.
\subsection{Different LoRA Ranks} \label{sub-appx: lora-ranks}
\begin{figure}[h]
  \centering
  \includegraphics[width=\linewidth]{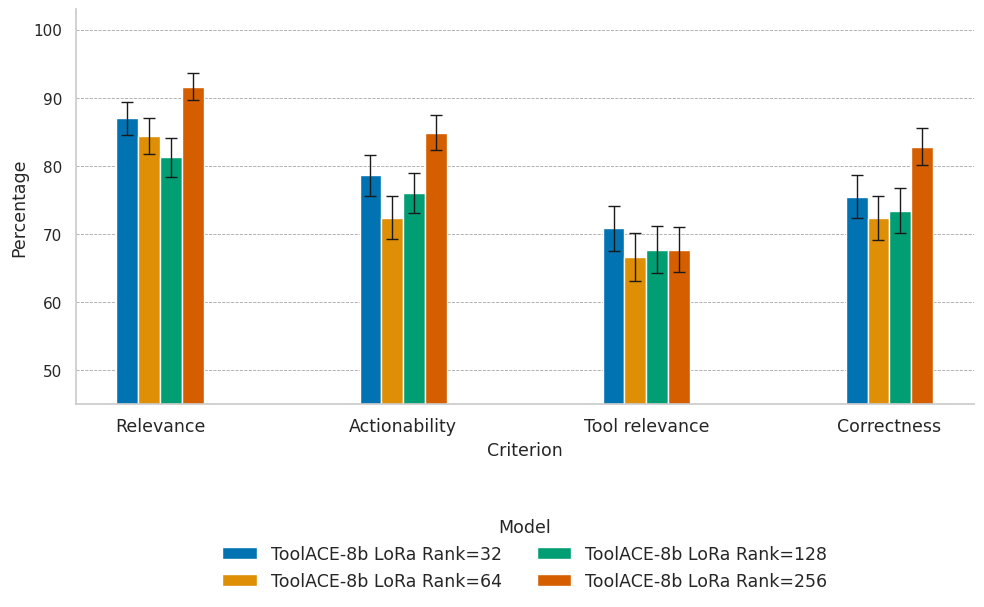}
  \caption{Percentage of YES given by the GPT-as-Judge for each criterion on the 192 evaluation questions (GEO, DSP and VA) on different LoRA ranks for ToolACE-8B-SCRIBE. }
  \label{fig:toolace-ranks}
\end{figure}

\begin{figure}[h]
  \centering
  \includegraphics[width=\linewidth]{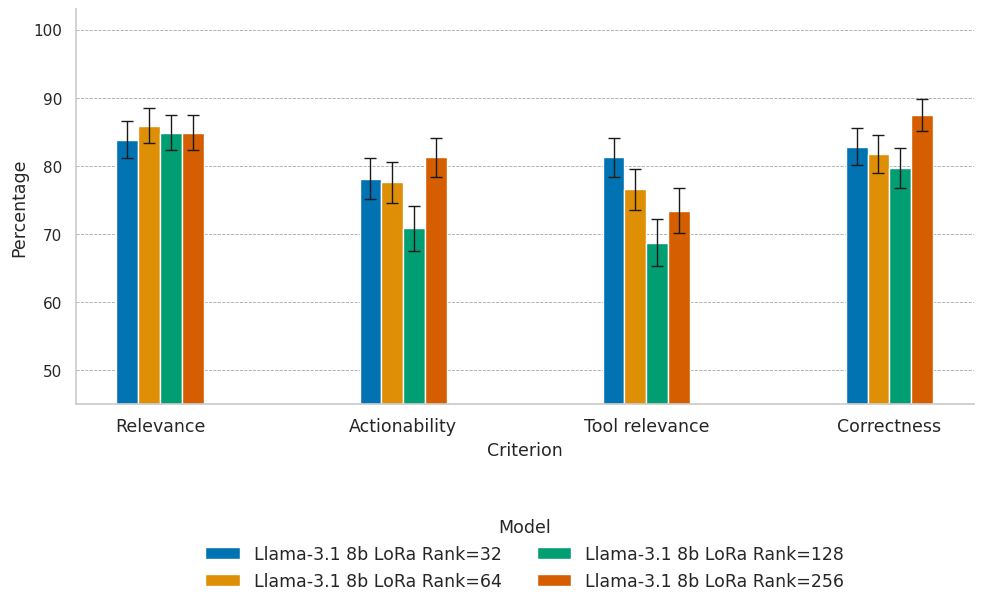}
  \caption{Percentage of YES given by the GPT-as-Judge for each criterion on the 192 evaluation questions (GEO, DSP and VA) on different LoRA ranks for Llama-3.1-8B-SCRIBE.}
  \label{fig:8B-ranks}
\end{figure}

In this section, we ablate the LoRA rank used for fine-tuning models on multihop reasoning with tool calling. As shown in \cref{fig:toolace-ranks,fig:8B-ranks}, we compare rank sizes 32, 64, 128, and 256 across both fine-tuning stages for the ToolACE-8B and Llama-3.1-8B models. Results indicate that rank 256 consistently outperforms lower ranks on actionability, and correctness for both models. It also out performs lower ranks on relevance in the case of ToolACE. An exception is tool relevance, where rank 32 achieves the highest performance. For Llama-3.1-8B, relevance is less sensitive to LoRA rank, but the model follows the same trend as ToolACE-8B on the other criteria.

\subsection{Single Stage vs two-stage LoRA} \label{sub-appx: single-vs-2-stage}

\begin{figure}[h]
  \centering
  \includegraphics[width=\linewidth]{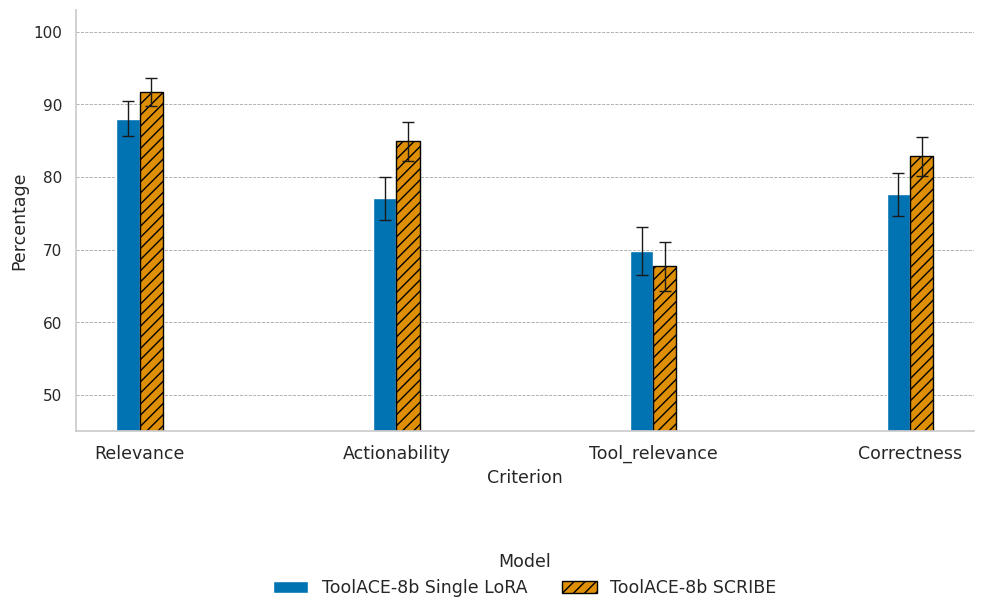}
  \caption{Percentage of YES given by the GPT-as-Judge for each criterion on the 192 evaluation questions (GEO, DSP and VA) to compare between single and multi stage LoRA}
  \label{fig:toolace-single-multi}
\end{figure}

\begin{figure}[h]
  \centering
  \includegraphics[width=\linewidth]{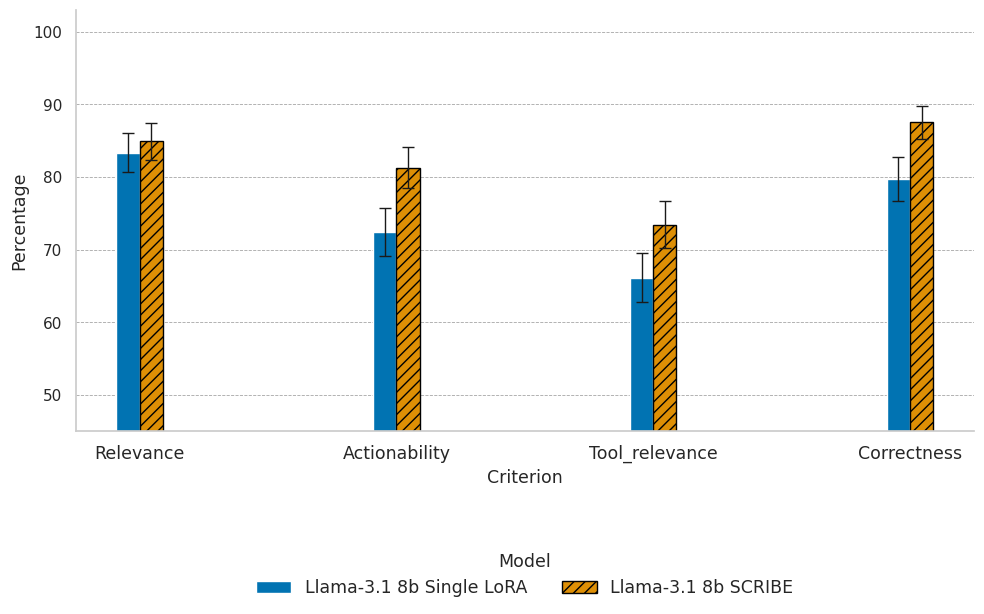}
  \caption{Percentage of YES given by the GPT-as-Judge for each criterion on the 192 evaluation questions (GEO, DSP and VA) to compare between single and multi stage LoRA}
  \label{fig:8B-single-multi}
\end{figure}

We additionally ablate our two-stage LoRA approach versus single LoRA in which the model was finetuned on single, multi-hop tool calling and final response formulation in a single stage. \cref{fig:toolace-single-multi,fig:8B-single-multi} shows the comparison between the approaches for ToolACE-8B and LLama-3.1-8B models respectively. While the only exception is the tool relevance only for the ToolACE model where the two-stage is slightly less, the figures show the two-stage LoRA consistently outperform single LoRA finetuning across all evaluation criteria for both models. This highlights the effectiveness of our multi-stage LoRA finetuning technique.  

\section{GPT-as-a-Judge} \label{appx:GPT-as-a-Judge}
In this section, we report the rubric defined by the annotators for each evaluation criterion as well as the per-category alignment, and the prompt used with GPT-4.1 for evaluation.
\subsection{Evaluation Rubric}
In the following, we describe the rubric agreed upon by human annotators for the judge. We explain the criteria used for judging the final response in terms of relevance, actionability, tool relevance, spelling and grammar, and correctness respectively. 
\begin{tcolorbox}[
    colback=pink!20,
    colframe=black!75!white,
    sharp corners,
    boxrule=0.5pt,
    width=\columnwidth,
    title=Human Annotators Rubric -- Relevance,
    breakable,
    enhanced
]
The response from the model directly addresses the student’s question. If the answer includes relevant responses and also extraneous information, then the response is still YES. The answer doesn’t need to be very detailed to be considered relevant, as long as it meaningfully responds to the student's question. If the Response is vague, unrelated, or fails to address the core question, then the response is NO.
\end{tcolorbox}

\begin{tcolorbox}[
    colback=pink!20,
    colframe=black!75!white,
    sharp corners,
    boxrule=0.5pt,
    width=\columnwidth,
    title=Human Annotators Rubric -- Actionability,
    breakable,
    enhanced
]
The response provides clear steps or instructions for the student to take to answer their question. If there is no action that is relevant based on the question (the question is purely informational such as asking about course materials or grading), then the answer to this question is YES. If the response provides vague, unclear, or generic advice without actionable instructions, then mark it as NO. Fallback advice in case tools did not prove enough information counts as actionable if clear — provided it's not hallucination or made up information (it can be a summary of what the model got from the tools or feedback reports or general actionable advice that it doesn't contain specific details that need to be double checked with an external source).
\end{tcolorbox}

\begin{tcolorbox}[
    colback=pink!20,
    colframe=black!75!white,
    sharp corners,
    boxrule=0.5pt,
    width=\columnwidth,
    title=Human Annotators Rubric -- Tool Relevance,
    breakable,
    enhanced
]
The tools that the model called are conceptually relevant to answer the question and can produce a response that directly answers the student's question. If the model calls multiple tools, some of which produce errors, the answer is YES if one or more of the tools provide sufficient information to answer the question. Do not evaluate the accuracy of the tool output or the correctness of the information passed to the tools by the LLM in this step. Multiple tools can be equally relevant to the question. If the called tools can "in theory" sufficient to answer the questions without needing to call another follow up tool then mark as YES.
\end{tcolorbox}

\begin{tcolorbox}[
    colback=pink!20,
    colframe=black!75!white,
    sharp corners,
    boxrule=0.5pt,
    width=\columnwidth,
    title=Human Annotators Rubric -- Spelling and Grammar,
    breakable,
    enhanced
]
The response is understandable without grammatical mistakes.
\end{tcolorbox}

\begin{tcolorbox}[
    colback=pink!20,
    colframe=black!75!white,
    sharp corners,
    boxrule=0.5pt,
    width=\columnwidth,
    title=Human Annotators Rubric -- Correctness,
    breakable,
    enhanced
]
The response is factually correct and strictly aligns with the provided tool outputs and course feedback context without any extrapolations or assumptions beyond the given data (tool outputs and feedback reports). Comparing tool arguments and outputs to the LLM response can be crucial for an accurate evaluation. For instance, if the response mentions weeks 4 and 5, but the tool was only called with week 4 as an argument, then the LLM is extrapolating the tool output and should be marked as NO. Only penalise tool misuse if it affects the final answer, rendering it factually incorrect. It is okay if the model relies entirely on the feedback report to provide an answer. It is also okay if the model says I couldn't find enough information and provide general "correct" advice. This is better than "not" saying that it couldn't find enough information and start making up unsupported claims information.
\end{tcolorbox}
\subsection{Per-category Alignment}
We report the per-category Cohen’s $\kappa$ for alignment between human and GPT as well as between both human annotators in~\cref{tab:kappa-category}.

\begin{table}[ht]
\centering
\resizebox{\columnwidth}{!}{
\begin{tabular}{lcc}
\toprule
\textbf{Metric} & \textbf{Human-GPT} & \textbf{Human-Human} \\
\midrule
Relevance            & 0.861 ± 0.000     & 0.755 \\
Tool Relevance       & 0.775 ± 0.039     & 0.843 \\
Actionability        & 1.000 ± 0.000     & 1.000 \\
Correctness          & 0.814 ± 0.000     & 0.843 \\
Overall $\kappa$     & 0.818 ± 0.014     & 0.850 \\
\bottomrule
\end{tabular}}
\caption{Cohen’s $\kappa$ Scores between human annotations and GPT and both human annotators.}
\label{tab:kappa-category}
\end{table}
\subsection{Evaluation Prompts}
Using the rubric agreed upon by humans, we use the following prompt to GPT-4.1. For this prompt, we feed the criterion and reasoning for CoT prompting depending on the evaluation criterion. In the following, we show the general prompt followed by the specific CoT prompt used for every criterion. 

\begin{tcolorbox}[
    colframe=black!75!white,
    sharp corners,
    boxrule=0.5pt,
    width=\columnwidth,
    title=Prompt for Evaluation,
    breakable,
    enhanced
]
\textbf{You are an impartial AI Judge} evaluating the \texttt{\{criterion\}} of a response provided by an AI assistant to a student question about their feedback report. Evaluate this criterion systematically using the reasoning process provided below.

\section*{Provided Materials}
\begin{itemize}
    \item \textbf{Tools Available for the AI Assistant}: \texttt{\{tool\_schemas\}}
\end{itemize}

\section*{Evaluation Process for \texttt{\{criterion\}}}
\begin{enumerate}
    \item Restate the student's question in your own words.
    \item Summarize the AI assistant’s response.
    \item Summarize tool arguments used.
    \item Explain your step-by-step reasoning regarding the \texttt{\{criterion\}} based on the definition provided.
    \item Make a clear \textbf{YES} or \textbf{NO} decision, explicitly justified by your reasoning.
\end{enumerate}

\hrulefill

\subsection*{\texttt{\{criterion\}} Definition}
\texttt{\{criterion\_definition\}}

\subsection*{Reasoning Steps}
\texttt{\{criterion\_reasoning\}}

\hrulefill

Please provide your evaluation for the \texttt{\{criterion\}} criterion only.

\vspace{0.5em}
\textbf{FINAL DECISION: YES or NO}

\end{tcolorbox}

\begin{tcolorbox}[
    colframe=black!75!white,
    sharp corners,
    boxrule=0.5pt,
    width=\columnwidth,
    title=CoT Prompt -- Relevance,
    breakable,
    enhanced
]

\textbf{Definition:}
\begin{itemize}
    \item \textbf{YES}: Response directly addresses the student's explicit question. It may include extra context or background information, as long as the core question is still clearly answered. \textbf{Do not} evaluate whether the correct tool was used or whether the response is accurate. If the response is on-topic and attempts to answer the student's question, even if it cannot provide exact details due to missing information, mark \textbf{YES}.
    \item \textbf{NO}: Response is vague, off-topic, or does not engage with the core of the student's question. This includes generic advice that does not attempt to answer the actual question asked.
\end{itemize}

\textbf{Reasoning Steps:}
\begin{itemize}
    \item \textbf{Step 1:} What specifically is the student asking?
    \item \textbf{Step 2:} Does the response directly engage with and attempt to answer that question?
    \item \textbf{Step 3:} Even if partially detailed or if the information is limited, does the response stay on-topic and provide a meaningful attempt to respond to the student's explicit request?
    \item \textbf{Important:} Do \textbf{not} penalize for incorrect tool usage or inaccurate content — that is evaluated under \textit{Correctness}.
\end{itemize}

\end{tcolorbox}

\begin{tcolorbox}[
    colframe=black!75!white,
    sharp corners,
    boxrule=0.5pt,
    width=\columnwidth,
    title=CoT Prompt -- Actionability,
    breakable,
    enhanced
]
\textbf{Definition:}
\begin{itemize}
    \item \textbf{YES}: The response explicitly provides clear steps, recommendations, or directions that the student can \textbf{reasonably follow}. If the question is \textbf{informational} (e.g., about course structure, exercises, resources, definitions, or explanations), mark \textbf{YES} automatically without reviewing the response, as no actions are required.

    If tool outputs \textbf{limit} the ability to offer detailed steps (e.g., no access to specific problems or resources), still mark \textbf{YES} if the response provides the \textbf{most practical and targeted guidance possible}—such as pointing to relevant topics, review areas, or general strategies tied to the tool output or feedback context.

    \item \textbf{NO}: Mark \textbf{NO} if the response is \textbf{vague}—e.g., generic, non-directional advice like "study more," "improve your skills," or "engage better" \textbf{without specifying what to focus on} or how to proceed. Also mark \textbf{NO} if it uses unexplained terms (e.g., "improve competency\_anticipation") or suggests unclear, impractical, or disconnected actions.
\end{itemize}

\textbf{Reasoning Steps:}
\begin{itemize}
    \item \textbf{Step 1:} Determine if the student’s question requires actionable guidance or is purely informational. Questions about content, exercises, resources, or definitions do not need an actionable response (\textbf{MARK YES} by default).\\
    \textit{Note: Requests for extra exercises or additional resources are not actionable and default to YES.}

    \item \textbf{Step 2:} If actionable, check whether the response provides \textbf{clear, focused, and applicable} steps or recommendations, even if high-level (e.g., “focus on topics like DFT and DTFT”).

    \item \textbf{Step 3:} If tool output restricts detailed actions, assess whether the response still offers \textbf{practical next steps} based on what’s available (e.g., pointing to relevant topics or materials).

    \item \textbf{Step 4:} Mark \textbf{NO} if the response only gives broad encouragement without direction (e.g., “engage more”) or includes technical terms without explanation.

    \item \textbf{Step 5:} Overall, if the student can \textbf{clearly understand what to do next}—even generally—mark \textbf{YES}. Do not assess tool relevance, usefulness, or correctness here.
\end{itemize}

\end{tcolorbox}

\begin{tcolorbox}[
    colframe=black!75!white,
    sharp corners,
    boxrule=0.5pt,
    width=\columnwidth,
    title=CoT Prompt -- Tool Relevance,
    breakable,
    enhanced
]

\textbf{Definition:}
\begin{itemize}
    \item \textbf{YES}: At least one chosen tool is conceptually appropriate for the question \textbf{and} is among the available tools for producing a correct or personalized answer. It does not have to be the best tool—only reasonably capable of generating the type of answer the student needs. \textbf{Do NOT} evaluate how well the tool was used or its output—only whether it was a strong choice given the available tools.
    
    \item \textbf{NO}: Either no tool was conceptually suited to the question, or the assistant used a tool when a clearly better, more appropriate tool was available and should have been used instead. This includes cases where the tool used cannot provide the type of information requested—e.g., using behavioral tools alone when the student asks about course topics, study strategies, or learning materials.
\end{itemize}

\textbf{Reasoning Steps:}
\begin{itemize}
    \item \textbf{Step 1:} Identify the type of information needed to answer the student’s question: performance patterns, general advice, conceptual understanding, study materials, or strategies.
    
    \item \textbf{Step 2:} Identify which tools (from the available list, not just the ones used) are conceptually capable of providing that information.
    \begin{itemize}
        \item sort\_student\_features\_with
        \_importance is for behavioral/performance analysis and cannot support content explanations or study material suggestions.
        \item \texttt{get\_feature\_description} defines internal metrics and is not suited for topic or concept-level guidance.
        \item \textbf{Mark NO} if these tools are used \textbf{alone} for questions asking about course understanding, conceptual improvement, or finding resources.
    \end{itemize}

    \item \textbf{Step 3:} Determine if the assistant used a conceptually appropriate tool. If yes, mark \textbf{YES}. If a clearly mismatched tool was used—even if the answer sounds plausible—mark \textbf{NO}. Do not evaluate tool usage quality, arguments, or output.
\end{itemize}

\end{tcolorbox}

\begin{tcolorbox}[
    colframe=black!75!white,
    sharp corners,
    boxrule=0.5pt,
    width=\columnwidth,
    title=CoT Prompt -- Spelling and Grammar,
    breakable,
    enhanced
]

\textbf{Definition:}
\begin{itemize}
    \item \textbf{YES}: The response is clear, readable, and contains no major spelling or grammatical errors affecting comprehension. Minor errors are acceptable if they do not hinder understanding.
    \item \textbf{NO}: Errors significantly reduce readability or clarity.
\end{itemize}

\textbf{Reasoning Steps:}
\begin{itemize}
    \item \textbf{Step 1:} Check for any major grammar or spelling errors.
    \item \textbf{Step 2:} Decide if these errors significantly impact readability or clarity.
\end{itemize}

\end{tcolorbox}

\begin{tcolorbox}[
    colframe=black!75!white,
    sharp corners,
    boxrule=0.5pt,
    width=\columnwidth,
    title=CoT Prompt -- Correctness,
    breakable,
    enhanced
]

\subsection*{Correctness}

\textbf{Definition:}
\begin{itemize}
    \item \textbf{YES}: The response is factually correct, aligns with the provided tool outputs and course feedback context, and avoids unsupported or misleading claims. General strategies or logical assumptions are acceptable as \textbf{correct} interpretations of the tool (e.g., noting that low engagement may impact performance, if engagement is referenced). Phrases like “likely to be relevant” are acceptable. The response does \textbf{not} need to explicitly acknowledge missing information.
    
    \item \textbf{NO}: The response includes clear inaccuracies, misleading assumptions, or unjustified certainty not supported by tool outputs or feedback. This includes:
    \begin{itemize}
        \item Making definitive claims about unknowns (e.g., exact exam content without syllabus details).
        \item Incorrect tool usage (e.g., passing week numbers to tools requiring topic names). Accept course name variants (e.g., \texttt{dsp\_002} for \texttt{dsp}).
        \item Misinterpreting or misrepresenting tool outputs or feedback—e.g., inventing definitions or substituting meanings not supported by data.
        \item Any factual errors or distortions that could mislead or confuse the student.
    \end{itemize}
\end{itemize}

\textbf{Reasoning Steps:}
\begin{itemize}
    \item \textbf{Step 1:} Summarize the student’s question, tool outputs, tool arguments, and feedback reports.
    
    \item \textbf{Step 2:} Check for incorrect tool usage (e.g., wrong arguments). If present, mark \textbf{NO}.
    
    \item \textbf{Step 3:} Verify that each claim or recommendation is explicitly supported by tool outputs, feedback, or represents a \textbf{harmless, logical educational strategy}. Do \textbf{not} accept reinterpreted meanings or invented definitions. Pay close attention to topic names, weeks, tool metrics, or feature names. Misuse of these—even if plausible—should be marked \textbf{NO} if potentially misleading.
    
    \item \textbf{Step 4:} General advice (e.g., study tips) and harmless assumptions (e.g., “missing content may impact performance”) are allowed without tool support, \textbf{as long as they do not misinterpret or substitute tool meanings}. Phrases like “likely to help” are fine. Penalize only if the advice introduces harmful specifics or misleading certainty.
    
    \item \textbf{Step 5:} If unknown information is presented as certain (e.g., stating guaranteed exam content), mark \textbf{NO}.
    
    \item \textbf{Step 6:} Ensure there are no harmful extrapolations, misinterpretations, or misleading assumptions. Even if harmless, unsupported claims (e.g., made-up definitions) must be rejected. Suggestions like reviewing extra material are acceptable, but definitions or specific answers must come from tools or the feedback report. Do not penalize use of known details from the feedback report (e.g., preferences, course topics). Do \textbf{not} evaluate tool relevance or completeness—focus solely on factual alignment with tool outputs and feedback.
\end{itemize}

\end{tcolorbox}

\section{Student Questions Generation} \label{appx:data-gen}

\subsection{Questions generation prompt}  \label{appx:quest-gen-prompt}
To generate questions that are close to those written by students, we use persona-based prompting \cite{wang-etal-2024-unleashing,10.5555/3721041.3721046} with GPT-4o. Each prompt simulates the scenario students encountered during the data collection phase (see \cref{sec: human_data_collection}) and includes the MOOC name, the feedback report, the question category (What have I done well?, Where should I improve?, How should I improve?, and What should I do next time?) and a set of guidelines derived from student comments and preferences observed during the study. Note that all feedback reports used for generating the questions were in English, and all generated questions are also in English.


\definecolor{lightblack}{rgb}{0.85,0.92,1}  

\begin{tcolorbox}[
    colframe=black!75!white,
    sharp corners,
    boxrule=0.5pt,
    width=\columnwidth,
    title=Prompt for Question Generation,
    breakable,
    enhanced
]

You are a student taking the an Online Course (MOOC): \textbf{\{course\_name\}}. Since the courses are difficult, often with low passing rates, the teaching team wants to help students who are not doing well to perform better in the course by giving them personalized assistance, and encourage students who are already performing well to continue.

Our goal is to give students feedback on their performance and possible trajectories. To do this, we use various weekly behavior features (such as the number of video clicks or how accurately questions are answered on weekly quizzes). We predict student performance early in the course (before the halfway point) as passing or failing behavior. We use the explanation of the prediction to give students additional, personalized feedback to help pass the course.

You received the following \textbf{personalized feedback report: \{feedback\_report\}}

\vspace{0.5em}
\hrule
\vspace{0.5em}

\textbf{Your Task:}
\begin{itemize}
    \item Generate \textbf{follow-up questions} in the style: \textbf{\{style\}}, defined as \textbf{\{question\_styles[style]\}}.
    \item Sound like a student: use \textbf{simple}, informal language, include \textbf{grammatical mistakes}, \textbf{short, direct}, or incomplete questions.
    \item Refer to these student examples: \textbf{\{questions\_sample\}} (don't copy — generate new ones).
    \item Include:
    \begin{itemize}
        \item Short: \emph{"Why did my score drop?"}
        \item Medium: \emph{"How can I use Week 2 to help later weeks?"}
        \item Long: \emph{"Week 7 not in report, but says prep for 6 and 8. Does that mean Week 7 is easier?"}
    \end{itemize}
\end{itemize}

\vspace{0.5em}
\hrule
\vspace{0.5em}

\textbf{Guidelines for Generating Questions:}
\begin{enumerate}
    \item Use everyday student language. Typos and grammar issues are okay.
    \item Ask about specific actions: e.g., \emph{"Should I rewatch Week 5 videos?"}
    \item Keep questions direct and practical.
    \item Avoid abstract or overly technical questions.
    \item Do not ask about general habits or external resources.
    \item Show emotion or stress, e.g., \emph{"I did bad, what to fix?"}
    \item Focus on content: Week 5 priority, quizzes, misunderstood topics.
    \item Avoid overused questions like:
    \begin{itemize}
        \item \emph{"Why did my score drop?"}
        \item \emph{"What can I do to improve?"}
        \item \emph{"Week X wasn’t mentioned, why?"}
    \end{itemize}
    \columnbreak
    \item Long questions (40+ words) should involve improvement strategies or specific content, not scheduling.
\end{enumerate}

\end{tcolorbox}

\subsection{Generated Questions Analysis} \label{appx:quest-gen-analysis}
\begin{figure}[H]  
  \centering
  \begin{subfigure}[t]{0.32\linewidth}
    \includegraphics[width=\linewidth]{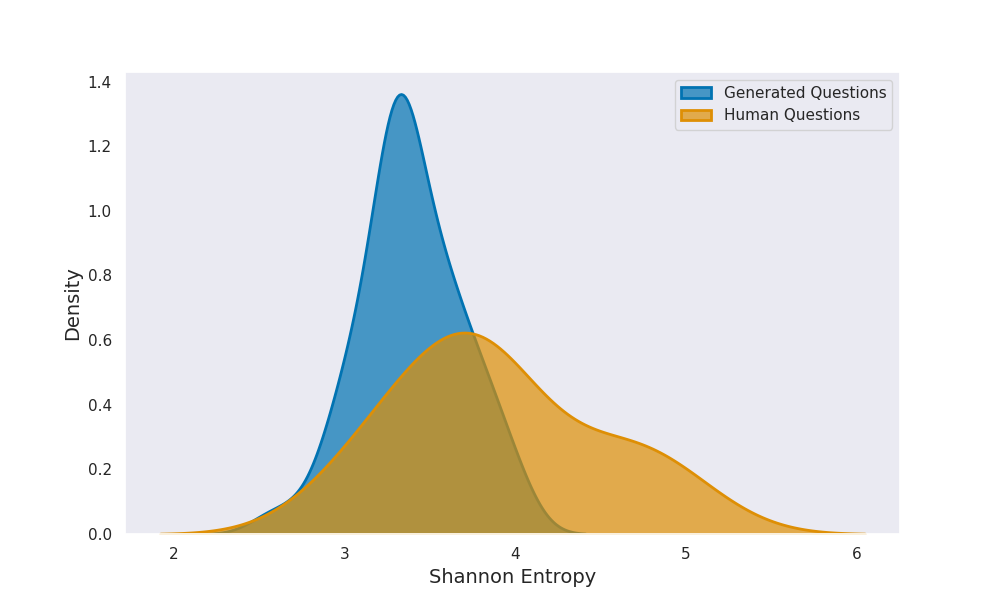}
    \caption{African Cities}
  \end{subfigure}
  \begin{subfigure}[t]{0.32\linewidth}
    \includegraphics[width=\linewidth]{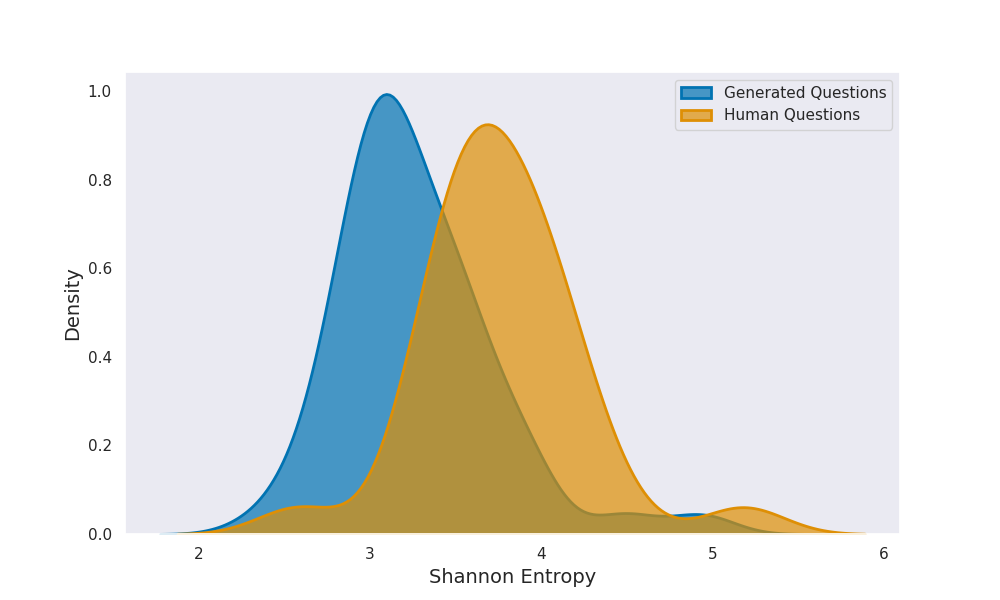}
    \caption{Digital Signal Processing}
  \end{subfigure}
  \begin{subfigure}[t]{0.32\linewidth}
    \includegraphics[width=\linewidth]{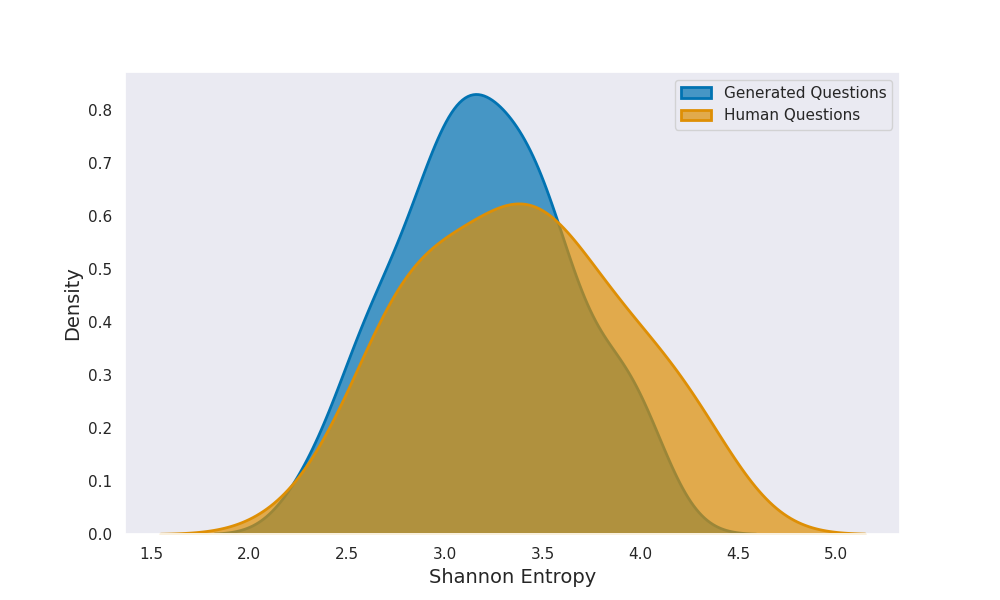}
    \caption{Elements of Geometry}
  \end{subfigure}
  \caption{KDE plots of Shannon Entropy for human vs. generated questions across MOOCs}
  \label{fig:entropy_kde_moocs}
\end{figure}

\begin{figure}[H]  
  \centering
  \begin{subfigure}[t]{0.32\linewidth}
    \includegraphics[width=\linewidth]{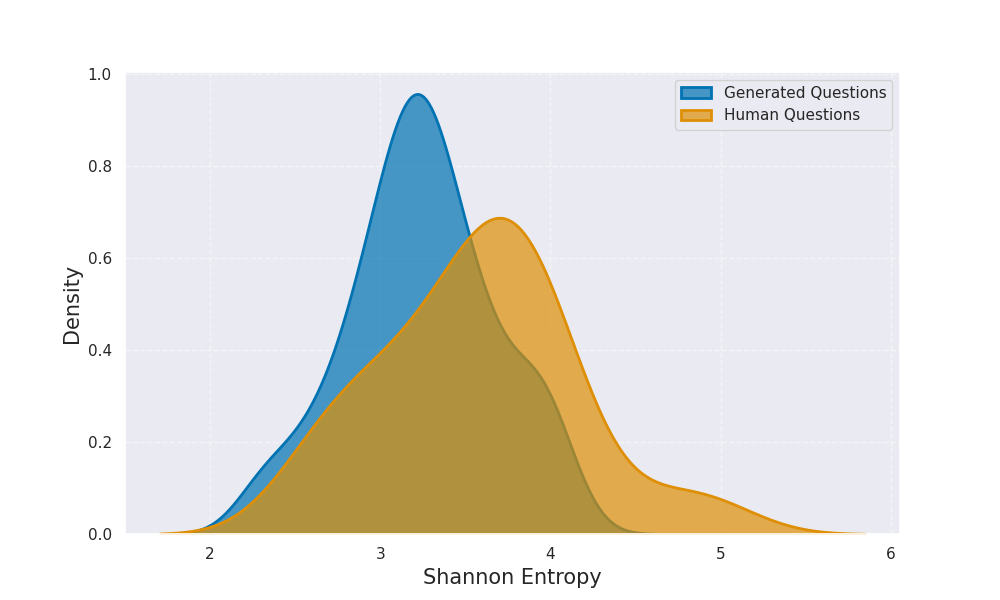}
    \caption{How can I improve?}
  \end{subfigure}
  \begin{subfigure}[t]{0.32\linewidth}
    \includegraphics[width=\linewidth]{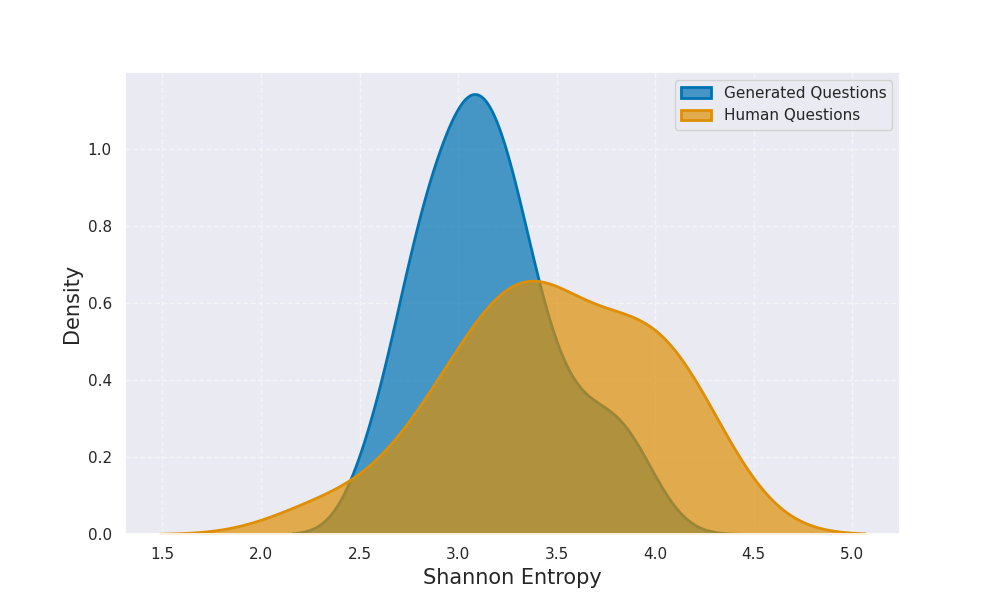}
    \caption{Where to improve?}
  \end{subfigure}
  \begin{subfigure}[t]{0.32\linewidth}
    \includegraphics[width=\linewidth]{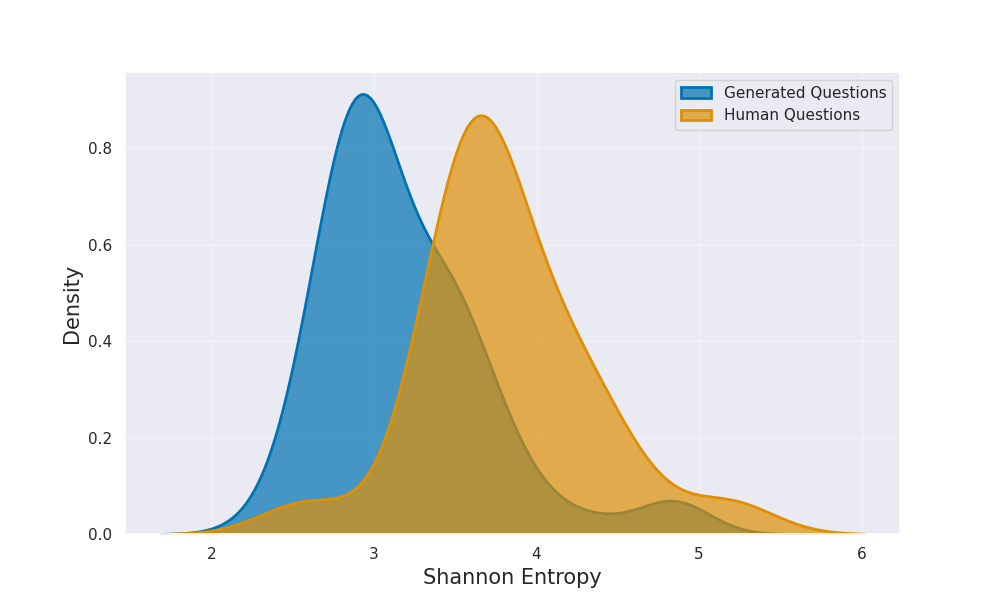}
    \caption{What to do next time?}
  \end{subfigure}
  \caption{KDE plots of Shannon Entropy across question types}
  \label{fig:entropy_kde_types}
\end{figure}

\begin{figure}[H]
  \centering
  \begin{subfigure}[t]{0.32\linewidth}
    \includegraphics[width=\linewidth]{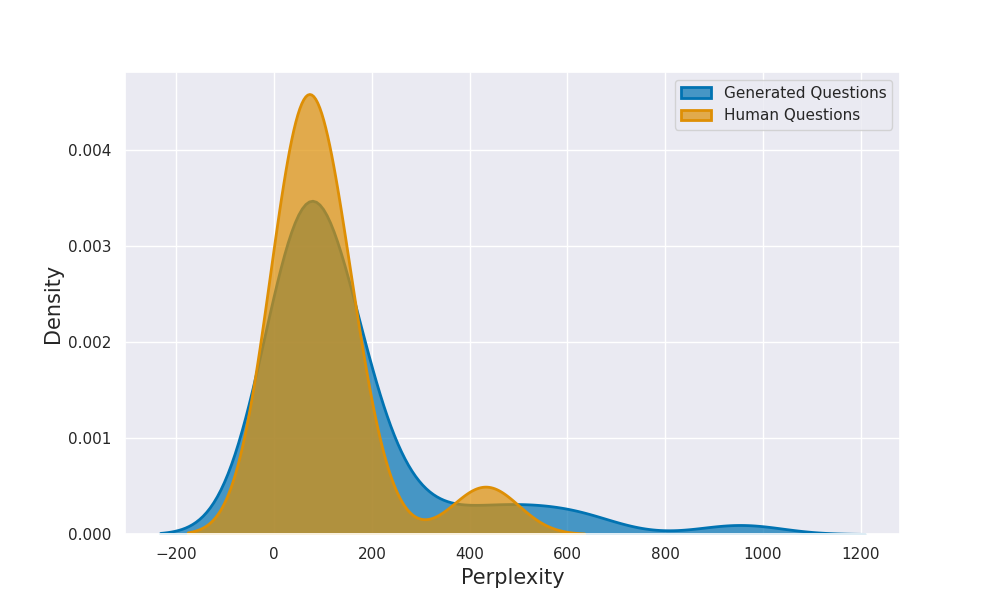}
    \caption{African Cities}
  \end{subfigure}
  \begin{subfigure}[t]{0.32\linewidth}
    \includegraphics[width=\linewidth]{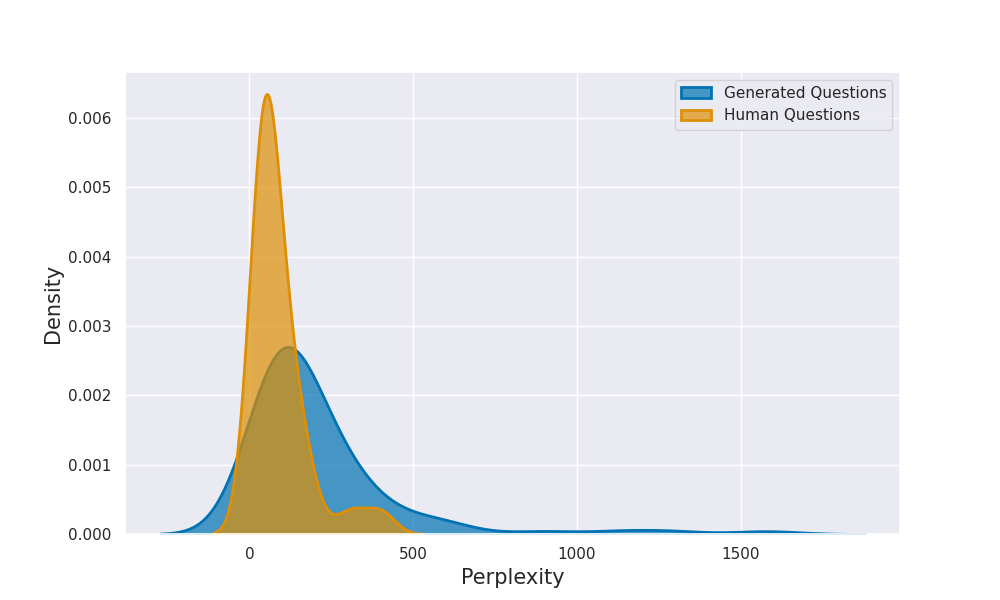}
    \caption{Digital Signal Processing}
  \end{subfigure}
  \begin{subfigure}[t]{0.32\linewidth}
    \includegraphics[width=\linewidth]{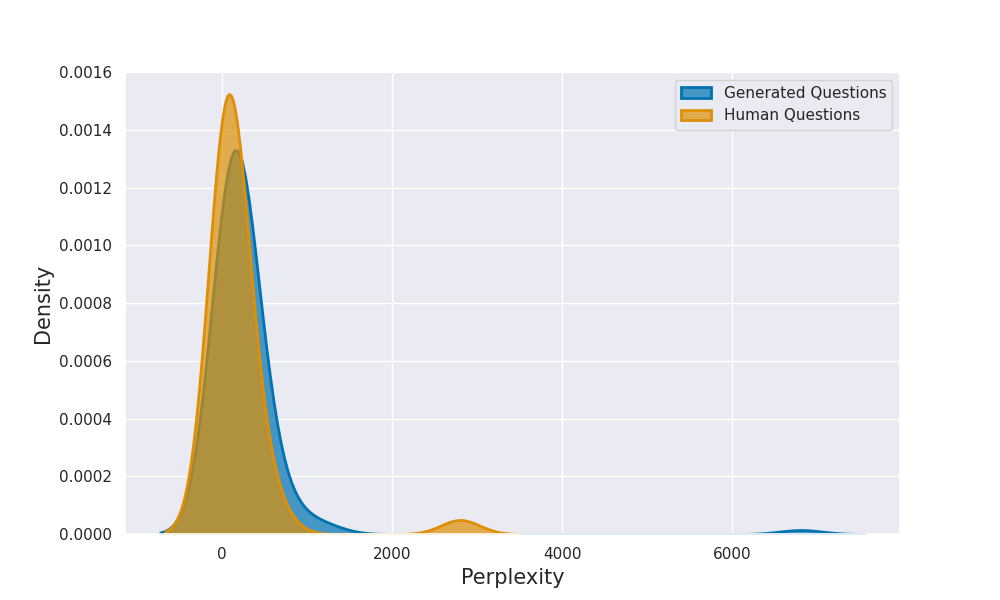}
    \caption{Elements of Geometry}
  \end{subfigure}
  \caption{KDE plots of Perplexity across MOOCs}
  \label{fig:perplexity_kde_moocs}
\end{figure}

\begin{figure}[H] 
  \centering
  \begin{subfigure}[t]{0.32\linewidth}
    \includegraphics[width=\linewidth]{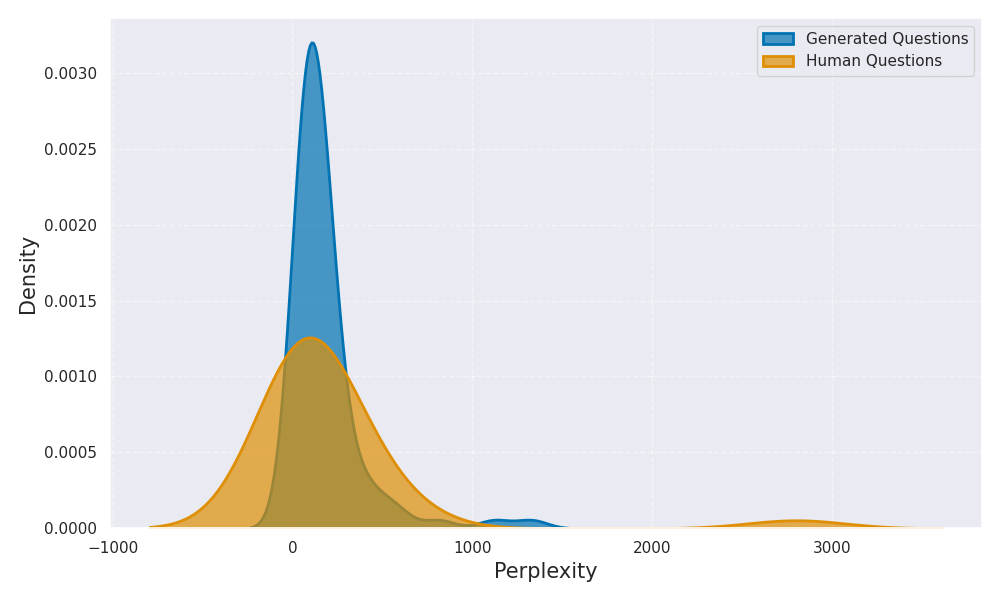}
    \caption{How can I improve?}
  \end{subfigure}
  \begin{subfigure}[t]{0.32\linewidth}
    \includegraphics[width=\linewidth]{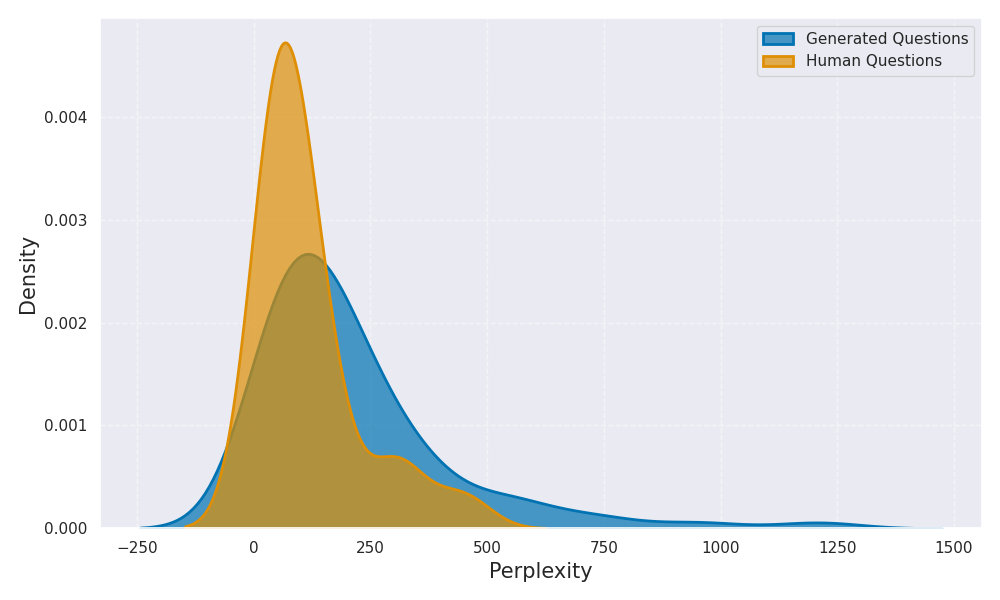}
    \caption{Where to improve?}
  \end{subfigure}
  \begin{subfigure}[t]{0.32\linewidth}
    \includegraphics[width=\linewidth]{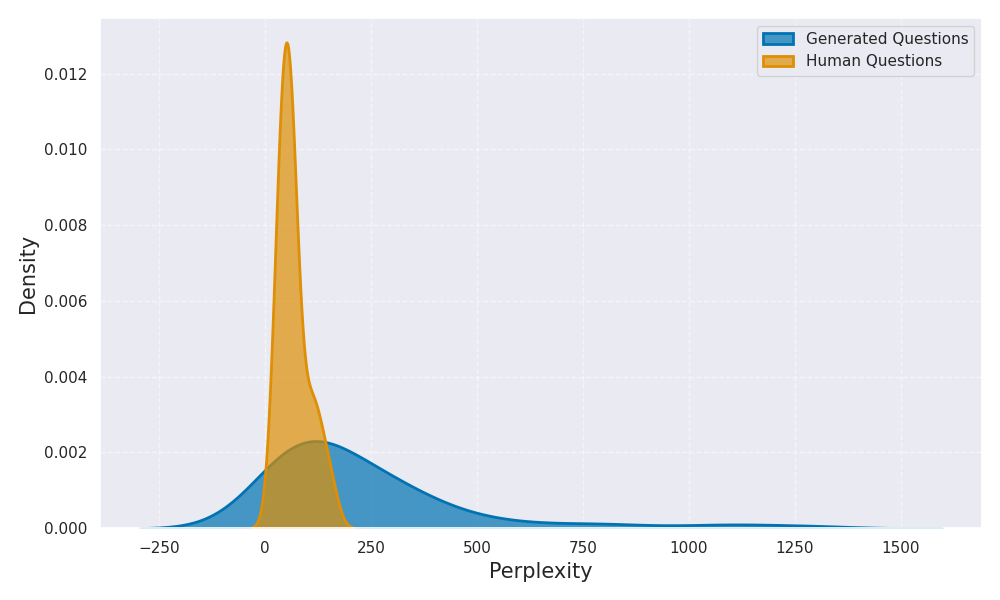}
    \caption{What to do next time?}
  \end{subfigure}
  \caption{KDE plots of Perplexity across question types}
  \label{fig:perplexity_kde_types}
\end{figure}

\begin{figure}[H]  
  \centering
  \begin{subfigure}[t]{0.45\linewidth}
    \includegraphics[width=\linewidth]{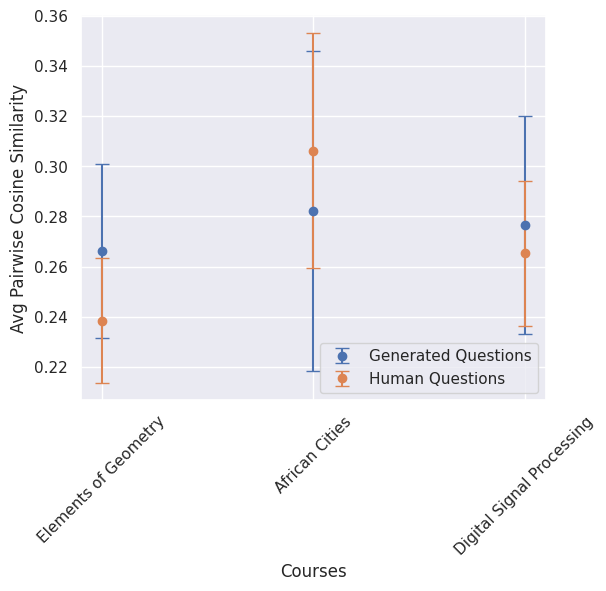}
    \caption{Pairwise Cosine Similarity by MOOC}
  \end{subfigure}
  \begin{subfigure}[t]{0.45\linewidth}
    \includegraphics[width=\linewidth]{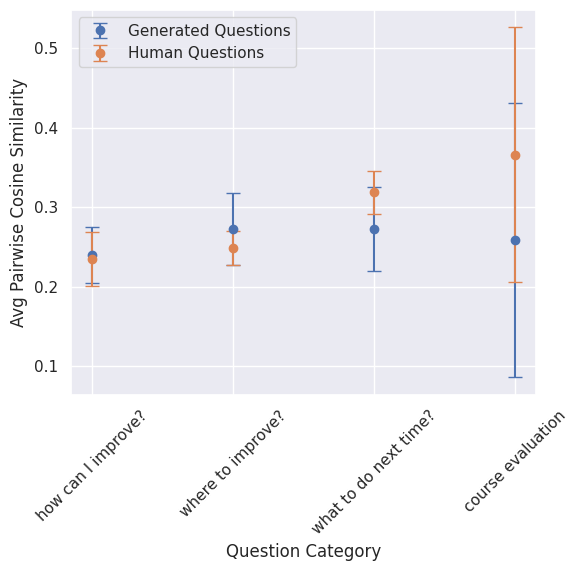}
    \caption{Pairwise Cosine Similarity by Feedback Category}
  \end{subfigure}
  \caption{Cosine similarity comparisons of real vs. generated questions}
  \label{fig:cosine_sim_questions}
\end{figure}

To compare real student questions with those generated by GPT-4o, we evaluate distributional similarity using Shannon entropy, perplexity, and cosine similarity. Figures \ref{fig:entropy_kde_moocs}–\ref{fig:cosine_sim_questions} show that generated questions closely match human-authored ones across feedback categories and MOOCs in terms of informativeness, fluency, and diversity.

\section{Tools: Topic Dependency Mapping.} \label{appx:tools}
In this section, we report the topic dependency maps created for the  Digital Signal
Processing (DSP) MOOC, the Elements de Géomatique (Geo) MOOC and the Villes Africaines (VA) MOOC used for the Topic Dependancy Mapping tool. Note that GEO and VA are taught in french while DSP and LNV are taught in English. We generate the VA and DSP maps in English and the GEO map in french.

\begin{figure}[h]
  \centering
  \includegraphics[width=\linewidth]{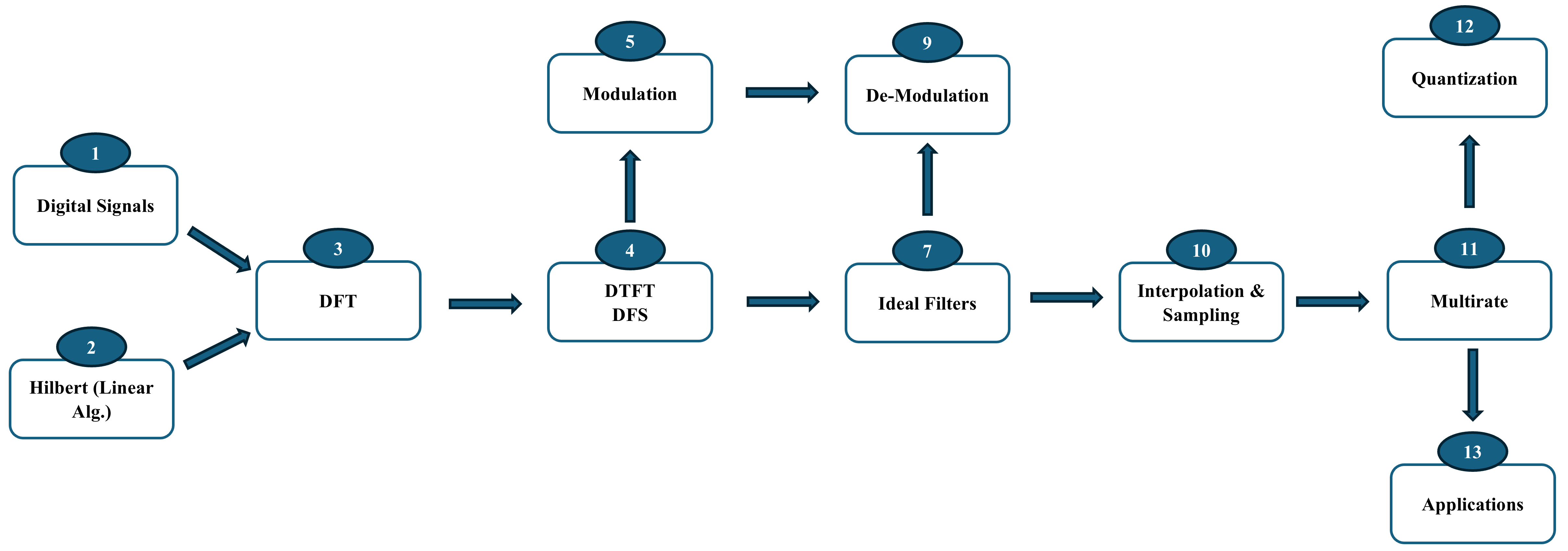}
  \caption{Digital Signal Processing (DSP) topic dependency map. The direction of each arrow indicates a dependency, where the source topic provides foundational knowledge required to understand the target topic}
  \label{fig:dsp depend map}
\end{figure}

\begin{figure}[h]
  \centering
  \includegraphics[width=\linewidth]{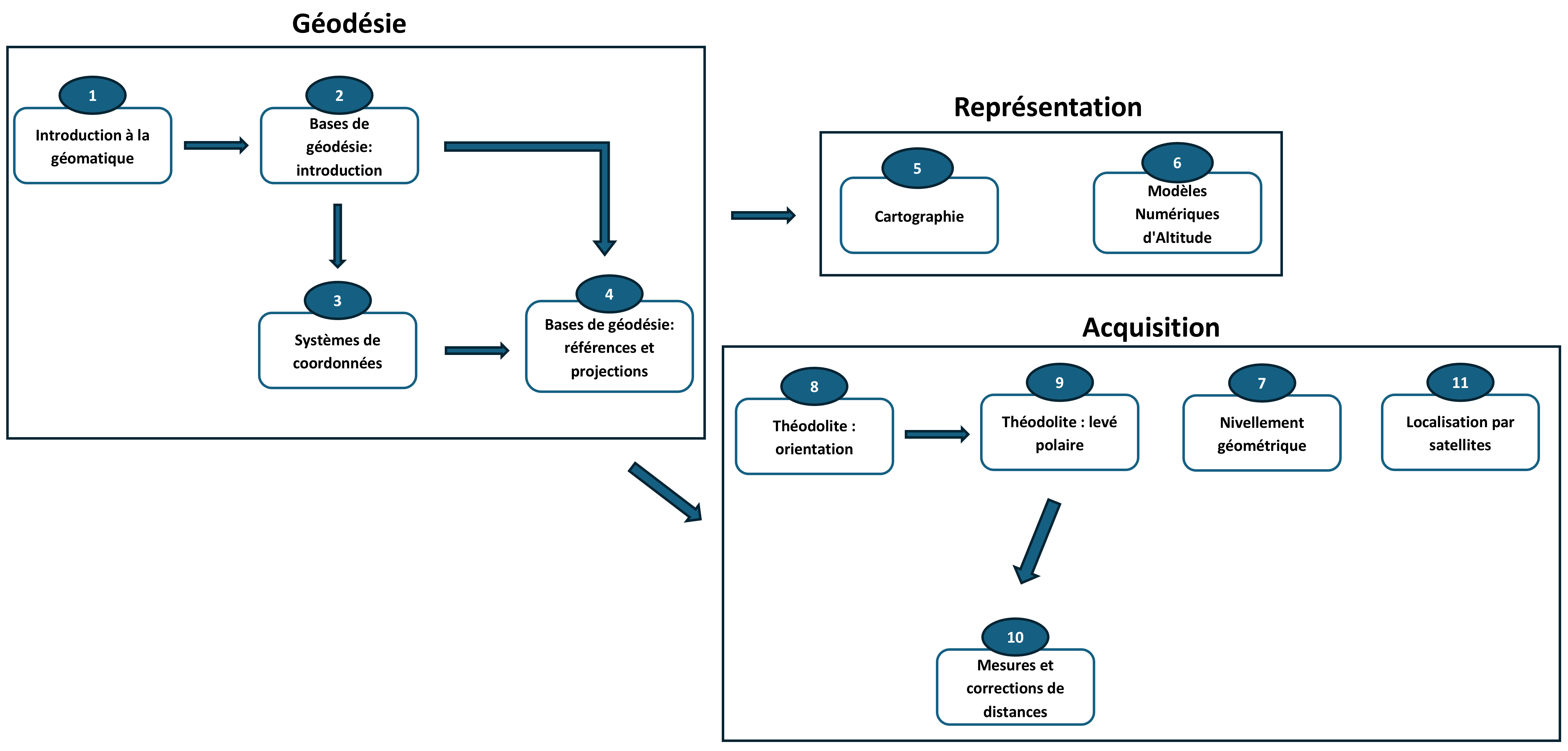}
  \caption{Elements de Géomatique (Geo)  topic dependency map. The direction of each arrow indicates a dependency, where the source topic provides foundational knowledge required to understand the target topic (or groups of topics)}
  \label{fig:geo depend map}
\end{figure}

\begin{figure}[h]
  \centering
  \includegraphics[width=\linewidth]{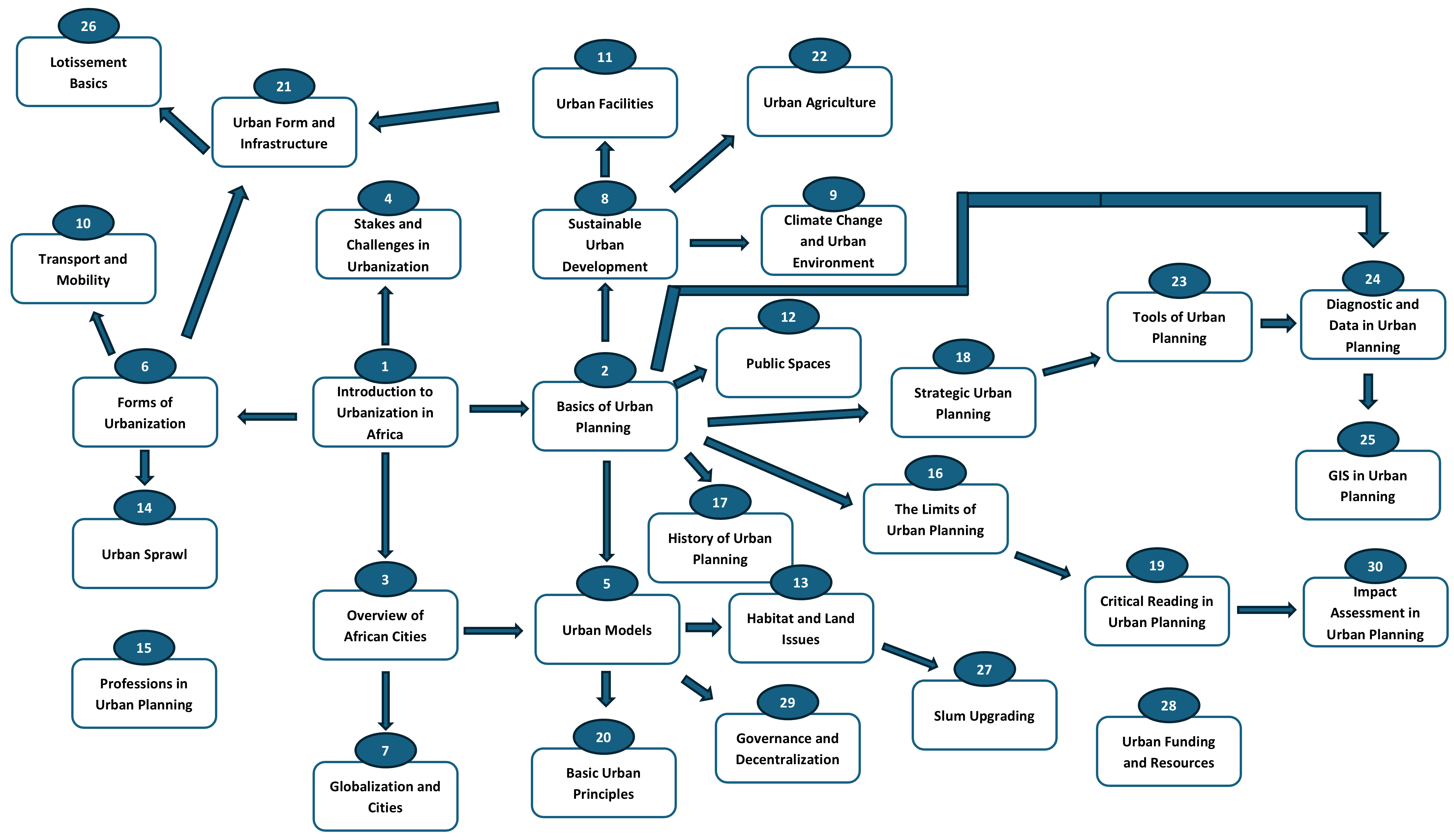}
  \caption{Villes Africaines (VA) topic dependency map. The direction of each arrow indicates a dependency, where the source topic provides foundational knowledge required to understand the target topic}
  \label{fig:VA depend map}
\end{figure}
\section{User Study} \label{appx:user-study}
In this section, we summarize the details of the user study we conducted. We start with details about the participants followed by the introduction used and ethical agreement. Finally, we show a statistical analysis of the user ratings results shown in~\cref{fig:user_study_rating}. 
\subsection{Participants Background}
\begin{figure}[!tbp]  
  \centering
  \includegraphics[width=\linewidth]{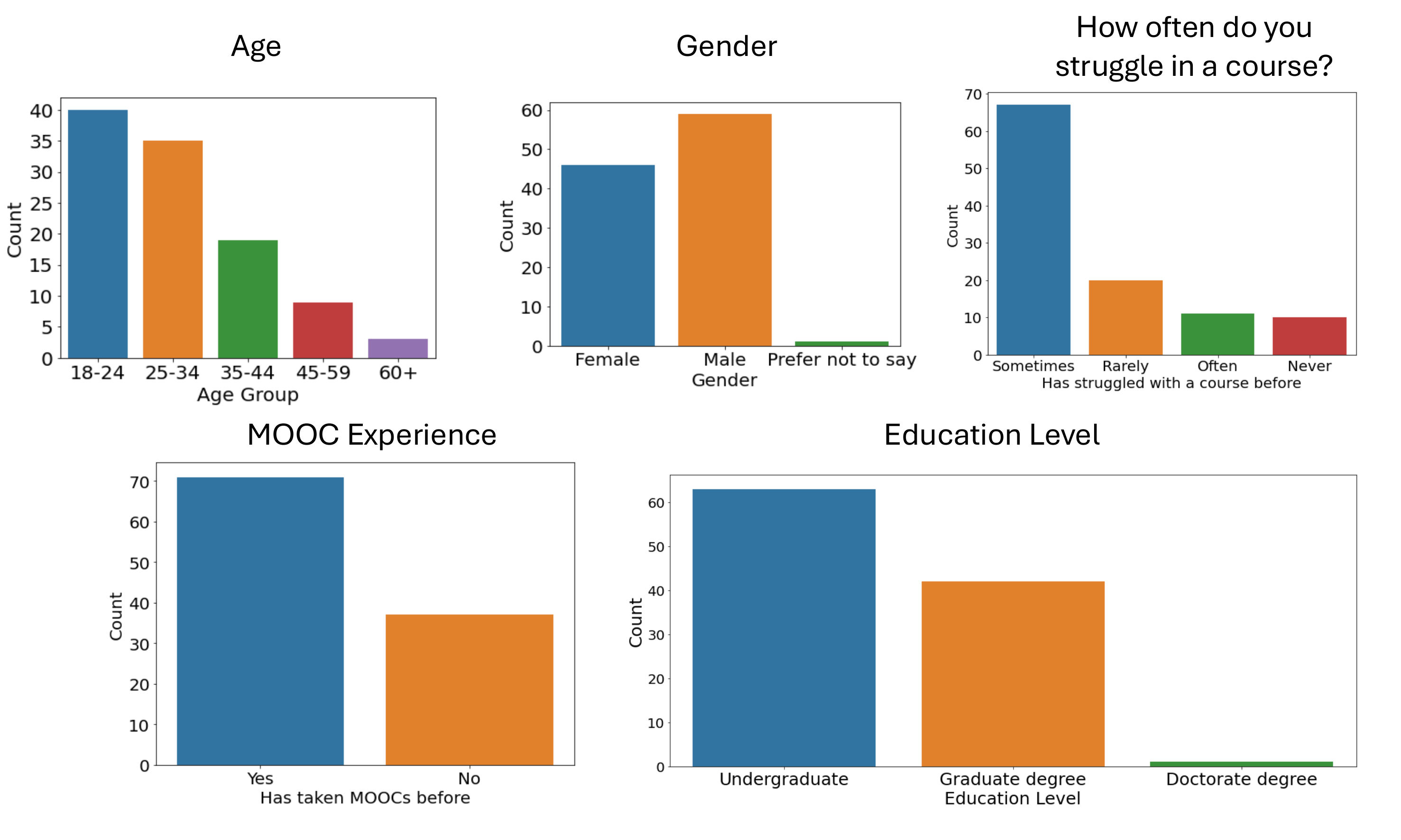}
  \caption{Demographics of study participants (age, gender, course struggles, MOOC Experience, and educational
background)}
  \label{fig:demographics}
  \vspace{-3 mm}
\end{figure}

We recruited 108 participants via Prolific, selecting individuals aged 18 and older who identified as students. As post-secondary students, they were well-positioned to engage with the academic context and assess the clarity and usefulness of the explanations provided. During the study, we gathered data on their experience with online courses (MOOCs), education level, and confidence in handling academic tasks (See \cref{fig:demographics} for the detailed demographics). The median completion time was 35 minutes, and participants earned an average hourly rate of £9.00 which was the recommended rate by the platform based on the participants' demographics. 

\subsection{User Study Introduction}
All participants gave informed consent; they could not proceed without first reading and agreeing to the explanatory statement in the introduction section of the study outlined below. 

\begin{tcolorbox}[
    colback=green!10,
    colframe=black!75!white,
    sharp corners,
    boxrule=0.5pt,
    width=\columnwidth,
    title=User Study Introduction Section,
    breakable,
    enhanced
]

Dear participant, 

Thank you for participating in our study on model explanations. We are very grateful for your participation and your invaluable insight. Please read this Explanatory Statement in full before proceeding. If you would like further information regarding any aspect of this project, please contact us using the email address provided below.

We are a group of researchers from the ML4ED Laboratory at EPFL, dedicated to improving education through technology. The goal of this study is to evaluate the responses of a language model when asked questions about progress feedback reports given to students to help improve their performance in an online course.

\subsection*{Human Research Ethics}

This survey has been approved by the EPFL Human Research Ethics Committee (HREC) under application number HREC 065-2022/27.09.2022. HREC reviews research proposals involving human participants to ensure they are ethically acceptable.

\begin{itemize}
    \item All personal information will be kept confidential and anonymized. Only demographic information is recorded, and it will be reported only in aggregate form to prevent identifying any individual participant.
    \item You may withdraw at any time. Any data you have provided up to that point will be destroyed.
    \item All data will be collected, stored securely, and reported in accordance with Swiss Federal law on data protection (Loi fédérale sur la protection des données – RS 235.1).
    \item Only anonymized or aggregated data may be used in future research (subject to ethics approval) and made available to other researchers for further analysis and verification.
    \item Only the principal investigator and the designated researchers will have access to the original data under strict confidentiality. Results from the project may be published in conference papers and/or journal articles, but no personal data will be shared.
    \item Personal data will be stored for 5 years from the date of collection. During this period, participants have the right to access their data and inquire about its processing. To exercise this right, contact the Principal Investigator.
\end{itemize}

By participating in this survey, you agree that your data may be used for scientific purposes.

In the following study, you will read \textbf{three progress feedback reports} and interact with a chatbot designed to answer your questions about each report. You will be expected to ask \textbf{three questions} per report. The study should take approximately 30 minutes. Please ensure you have sufficient time to complete it in full, as incomplete submissions will not be considered.

We ask that you approach the questions seriously and complete them to the best of your ability. Responses will be reviewed for quality, and submissions that appear unserious may be discarded. If you encounter any issues or would like to provide additional feedback or request more information, feel free to contact us.

\section*{Context}

You are a student enrolled in three online courses (MOOCs): \textit{Digital Signal Processing}, \textit{Elements of Geometry}, and \textit{Launching New Ventures}. These courses are known for their challenging content and typically low passing rates. To better support students, the teaching team has implemented a system that provides personalized feedback based on each student's learning behavior.

We used a highly accurate predictive model (over 90\% accuracy) to forecast student success or risk of failure early in the course, using weekly behavioral data (e.g., number of video views, quiz performance, engagement metrics). Based on these predictions, each student received a personalized feedback report explaining the factors influencing their predicted performance and offering tailored advice to improve or maintain success.

This study explores how students can interact with these feedback reports using a language model assistant. This assistant allows you to ask questions about your feedback report, clarify details, seek advice, and better understand the factors affecting your learning progress. To ensure accuracy, the assistant uses deterministic tools to retrieve precise information needed to answer your questions.

You will receive \textbf{three feedback reports} and are expected to ask \textbf{three to five clarifying questions} for each report. Questions must focus only on the feedback content. For the same report, you may ask different questions or a sequence of follow-ups.

\subsection*{Evaluation Criteria}

We will assess the assistant’s responses based on the following criteria:

\begin{itemize}
    \item \textbf{Relevance}: The response directly addresses the question without veering off-topic.
    \item \textbf{Usefulness}: The response provides meaningful insights that enhance learning or deepen understanding.
    \item \textbf{Actionability}: The response includes clear, practical steps or guidance relevant to the question.
    \item \textbf{Coverage}: The response thoroughly addresses all parts of the question, including sub-questions.
    \item \textbf{Conciseness}: The response is clear and complete, using the fewest words necessary while avoiding repetition or unnecessary detail.
\end{itemize}

\end{tcolorbox}

\subsection{User Study Ratings Analysis} \label{appx:anova}
Table \ref{tab:anova_userstudy} shows results of ANOVA test. For all criteria, we failed to reject the null hypothesis ($p > 0.05$), indicating no significant difference in perceived response quality.

\begin{table}[h]
\centering
\caption{One-way ANOVA comparing average ratings across models for each evaluation criterion. All $p$-values > 0.05 indicate no statistically significant difference.}
\resizebox{\columnwidth}{!}{
\begin{tabular}{lccccc}
\toprule
\textbf{} & \textbf{Actionability} & \textbf{Conciseness} & \textbf{Coverage} & \textbf{Relevance} & \textbf{Usefulness} \\
\midrule
\textbf{F-value} & 0.204 & 0.366 & 0.619 & 0.408 & 0.061 \\
\textbf{Degrees of freedom} & (2, 321) & (2, 321) & (2, 321) & (2, 321) & (2, 321) \\
\textbf{p-value} & 0.816 & 0.694 & 0.539 & 0.665 & 0.941 \\
\bottomrule
\end{tabular}}
\label{tab:anova_userstudy}
\end{table}
\section{Question Category Annotation Rubric} \label{appx:anno_rubric}

In this section, we provide the rubric used to categorize the question categories. They are adapted from \cite{hattie}.

\begin{tcolorbox}[
    colback=pink!20,
    colframe=black!75!white,
    sharp corners,
    boxrule=0.5pt,
    width=\columnwidth,
    title=Question Category Annotation Rubric -- how can I improve?,
    breakable,
    enhanced
]
"How to improve?” relates to how to correct certain errors or what strategies students can follow to rectify their problems. It should be related to current progress and how to fix current issues. Example: How can I do better in the weeks 3,4,5? 
\end{tcolorbox}

\begin{tcolorbox}[
    colback=pink!20,
    colframe=black!75!white,
    sharp corners,
    boxrule=0.5pt,
    width=\columnwidth,
    title=Question Category Annotation Rubric -- where to improve?,
    breakable,
    enhanced
]
“Where to improve?” Indicates where errors have occurred, and what needs to be fixed. This category includes questions that ask for elaboration on specific tasks or weaknesses in certain weeks or topics. Example: Why did my performance drop?
\end{tcolorbox}

\begin{tcolorbox}[
    colback=pink!20,
    colframe=black!75!white,
    sharp corners,
    boxrule=0.5pt,
    width=\columnwidth,
    title=Question Category Annotation Rubric -- what to do next time?,
    breakable,
    enhanced
]
“What to do next time?” relates to future directions, events or tasks that will be carried out in the future. This also encompasses self-regulation or questions regarding developing the capacity to self-monitor. Example: What is the best way to start reviewing for the next week's material? 
\end{tcolorbox}

\begin{tcolorbox}[
    colback=pink!20,
    colframe=black!75!white,
    sharp corners,
    boxrule=0.5pt,
    width=\columnwidth,
    title=Question Category Annotation Rubric -- course evaluation,
    breakable,
    enhanced
]
Relates to course evaluation criteria and non-improvement or feedback questions. Example: How is the evaluation of the course done? 
\end{tcolorbox}
\section{Inference Prompts} \label{appx:self-reflection-prompt}
We report an example of the self-reflection prompt used to correct errors in tool calling. We additionally provide prompts used for inference for the initial reasoning stage and the multiple reasoning stages. 
\begin{tcolorbox}[
    colback=lightblack,
    colframe=black!75!white,
    sharp corners,
    boxrule=0.5pt,
    width=\columnwidth,
    title=Self Reflection Prompt Example for Error Correction,
    breakable,
    enhanced
]

You encountered an error during reasoning or tool invocation.

\section*{Error Message}
I encountered an error: \texttt{\{str(e)\}}. Please fix your reasoning or calls so we can reach a final answer. \\
Remember to use the correct tokens for tool call and final answer: \texttt{[TOOL\_CALL]} and \texttt{[FINAL\_ANSWER]}. \\
Terminate them using: \texttt{[END\_OF\_TOOL\_CALL]} and \texttt{[END\_OF\_FINAL\_ANSWER]}. \\
\textbf{Note:} Without \texttt{[END\_OF\_TOOL\_CALL]} and \texttt{[END\_OF\_FINAL\_ANSWER]}, your answer cannot be parsed.


\medskip

\noindent
\texttt{
\string<|start\_header\_id|>\,user \\ 
\string<|end\_header\_id|> \\
{[}ERROR\_NOTICE{]}\{error\_message\}
\\ {[}/ERROR\_NOTICE{]} \\
\string<|eot\_id|>\string<|start\_header\_id|>\, \\
assistant\string<|end\_header\_id|> \\
{[}REASONING{]}
}

\end{tcolorbox}

\begin{tcolorbox}[
    colback=lightblack,
    colframe=black!75!white,
    sharp corners,
    boxrule=0.5pt,
    width=\columnwidth,
    title=Initial Stage Prompt,
    breakable,
    enhanced
]

You are a \textbf{reasoning tool-calling agent} tasked with \textbf{analyzing} a student's question about the personalized feedback they received. Students are enrolled in MOOC courses and have received individualized feedback on their learning progress and performance.

You do not know anything about the MOOCs or the student and are not allowed to give any advice or information that is not in the feedback report or the tool outputs.

\section*{Context}
\begin{itemize}[leftmargin=*]
    \item \textbf{Course Name}: \texttt{\{course\_name\}}
    \item \textbf{Student Feedback Report}: \texttt{\{feedback\_report\}}
\end{itemize}

\section*{Available Tools}
\texttt{\{tool\_schemas\}}

\section*{Your Task}
\begin{itemize}[leftmargin=*]
    \item Analyze the student's question in relation to their feedback report.
    \item Think about the best tool to use to answer the student's question.
    \begin{itemize}
        \item Use tools for behavior analysis when the question is about the student's behavior.
        \item Use \texttt{impact\_of\_student\_behaviors} for hypothetical or general behavioral questions (like time management, catching up, or study strategies). It does not provide personalized information about the student's specific activity.
        \item Use tools for course content when the question is about the course content.
        \item Use tools for course evaluation when the question is about the course evaluation.
        \item Use tools for student performance when the question is about the student's performance.
    \end{itemize}
    \item Provide a \textbf{reasoning} to determine the \textbf{first tool} needed to answer the student's question. Wrap your reasoning in \texttt{[REASONING]} and \texttt{[END\_OF\_REASONING]} tokens.
    \item Determine the \textbf{single best tool} from the tools above to retrieve that information.
\end{itemize}

\end{tcolorbox}

\begin{tcolorbox}[
    colback=lightblack,
    colframe=black!75!white,
    sharp corners,
    boxrule=0.5pt,
    width=\columnwidth,
    title=Multi Stage Prompt,
    breakable,
    enhanced
]
You are a \textbf{reasoning tool-calling agent} talking to a student and responsible for analyzing the student's question in relation to their personalized feedback. Students are enrolled in MOOC courses and receive individualized feedback on their learning progress and performance.

You will be talking to the student and you need to provide them with the best answer possible.

You do not know anything about the MOOCs or the student and are not allowed to give any advice or information that is not in the feedback report or the tool outputs.

\section*{Context}
\begin{itemize}[leftmargin=*]
    \item \textbf{Course Name}: \texttt{\{course\_name\}}
    \item \textbf{Student Feedback Report}: \texttt{\{feedback\_report\}}
\end{itemize}

\section*{Available Tools}
\texttt{\{tool\_schemas\}}

\section*{Task}
\begin{itemize}[leftmargin=*]
    \item Given the \textbf{student's question, previous reasoning, tool calls, and tool outputs}, determine whether another tool call is needed or if a final answer can be provided.
    \item If a tool call is needed:
    \begin{itemize}
        \item \textbf{Explain why} the tool call is required.
        \item \textbf{Generate the structured tool call.}
    \end{itemize}
    \item If the final answer can be provided:
    \begin{itemize}
        \item \textbf{Explain why} no further tool calls are needed.
        \item \textbf{Generate the structured final answer.}
    \end{itemize}
\end{itemize}

\section*{Response Format}
\begin{itemize}[leftmargin=*]
    \item \textbf{Always} wrap reasoning in \texttt{[REASONING]} ... \texttt{[END\_OF\_REASONING]}.
    \item \textbf{If making a tool call,} follow reasoning with \texttt{[TOOL\_CALL]} ... \texttt{[END\_OF\_TOOL\_CALL]}.
    \item \textbf{If providing the final answer,} follow reasoning with \texttt{[FINAL\_ANSWER]} ... \texttt{[END\_OF\_FINAL\_ANSWER]}.
    \item \textbf{Stop after the tool call or final answer.} Do not generate tool outputs or explanations beyond the required response.
    \item Do not use your own knowledge, only use the feedback report and the tool schemas.
\end{itemize}

\end{tcolorbox}

\section{Tool contribution analysis} \label{appx:tool_contributions}

To understand how the original toolset contributed during inference, we examined usage frequencies of all tools in ToolACE-8B SCRIBE on the DSP, GEO, and VA evaluation dataset. 
Table~\ref{tab:tool_usage} reports the percentage of calls per tool. 
Every tool was invoked at least once, with the most frequently used being \texttt{map\_week\_to\_topic} (28.84\%) and \texttt{impact\_of\_student\_behaviors} (26.22\%). 
Other tools were used less often but still contributed, indicating that the model relied on the full set of tools when responding to student questions. 

\begin{table}[htbp]
\centering
\resizebox{\columnwidth}{!}{%
\begin{tabular}{lc}
\toprule
\textbf{Tool} & \textbf{Percentage Use (\%)} \\
\midrule
dsp\_textbook\_exercise\_search & 0.75 \\
get\_course\_syllabus & 7.12 \\
get\_dependant\_topics & 14.23 \\
get\_feature\_description & 5.99 \\
grade\_calculator & 1.12 \\
impact\_of\_student\_behaviors & 26.22 \\
map\_week\_to\_topic & 28.84 \\
sort\_student\_features\_with\_importance & 12.73 \\
textbook\_search & 3.00 \\
\bottomrule
\end{tabular}%
}
\caption{Tool usage patterns for ToolACE-8B SCRIBE on the DSP, GEO, and VA evaluation dataset.}
\label{tab:tool_usage}
\end{table}

\end{document}